\RequirePackage{fix-cm}
\documentclass[11pt]{article}
\ifdefined\XeTeXversion\else\pdfoutput=1\fi

\usepackage[T1]{fontenc}

\usepackage{microtype}
\usepackage{graphicx}
\usepackage{float}
\usepackage{algorithm}
\usepackage{algpseudocode}
\usepackage{booktabs}
\usepackage{amssymb}
\usepackage{bm}
\usepackage{bbm}
\usepackage{url}
\usepackage{amsmath}
\usepackage{amsthm}
\usepackage{longtable}
\usepackage[table]{xcolor}
\usepackage{soul}
\usepackage{textcomp}
\usepackage{multirow}
\usepackage{mathtools}
\usepackage[margin=1in]{geometry}
\usepackage{authblk}
\usepackage[numbers]{natbib}
\usepackage{xspace}
\usepackage{array}
\usepackage{pifont}
\usepackage{subcaption}
\usepackage{titletoc}
\usepackage{comment}
\usepackage{wrapfig}
\usepackage{placeins}
\usepackage{thmtools,thm-restate}
\usepackage{verbatim}
\usepackage{enumitem}
\usepackage{tabularx}
\usepackage{caption}
\floatname{algorithm}{Algorithm}
\usepackage{pgfplots}\pgfplotsset{compat=1.18}
\usepackage{needspace}
\usepackage{fontawesome5}
\usepackage{tikz}
\usepackage[most]{tcolorbox}
\usetikzlibrary{arrows}
\usetikzlibrary{positioning}

\definecolor{linkblue}{RGB}{38,110,165}
\definecolor{linkpink}{HTML}{EC008C}

\definecolor{ncmgold}{HTML}{7A5200}
\definecolor{cardinal}{HTML}{820000}
\definecolor{cNCM}{HTML}{EDA100}
\definecolor{cNCMtxt}{HTML}{7A5200}
\definecolor{cNCMlight}{HTML}{FFE3A0}
\definecolor{cNCMlevel1}{HTML}{B87A00}
\definecolor{cNCMlevel2}{HTML}{E09A0E}
\definecolor{cNCMlevel3}{HTML}{F5BA3E}
\definecolor{cNCMlevel4}{HTML}{FAD48A}
\definecolor{cNeutral}{HTML}{8C8C8C}
\definecolor{cPrompt}{HTML}{5578A5}
\definecolor{cMDLM}{HTML}{008300}
\definecolor{cMDLMtxt}{HTML}{008300}
\definecolor{cDUO}{HTML}{4A3AA7}
\definecolor{cDUOtxt}{HTML}{4A3AA7}
\definecolor{cFLM}{HTML}{2A78D6}
\definecolor{cFLMtxt}{HTML}{2876D4}
\definecolor{cCANDI}{HTML}{E34948}
\definecolor{cCANDItxt}{HTML}{D63D3E}
\definecolor{cAURORALM}{HTML}{E87BA4}
\definecolor{cAURORALMtxt}{HTML}{BB537D}
\definecolor{cELF}{HTML}{1BAF7A}
\definecolor{cELFtxt}{HTML}{0A7A8A}
\definecolor{cEsoLM}{HTML}{E87BA4}
\definecolor{cEsoLMtxt}{HTML}{BB537D}
\definecolor{cSEDD}{HTML}{6C7A89}
\definecolor{cSEDDtxt}{HTML}{697786}
\definecolor{cRDLM}{HTML}{8C8C8C}
\definecolor{cRDLMtxt}{HTML}{767676}
\definecolor{cE2D2}{HTML}{4D4D4D}
\definecolor{cE2D2txt}{HTML}{4D4D4D}
\definecolor{cAR}{HTML}{6F6F6F}
\definecolor{cARtxt}{HTML}{6F6F6F}
\definecolor{cSFLM}{HTML}{2A78D6}
\definecolor{cSFLMtxt}{HTML}{2876D4}
\definecolor{cFMLM}{HTML}{882255}
\definecolor{cFMLMtxt}{HTML}{882255}

\definecolor{ncmLogoA}{HTML}{E09A0E}
\definecolor{ncmLogoB}{HTML}{F5BA3E}
\definecolor{ncmLogoC}{HTML}{FCE2A6}
\definecolor{ncmLogoEdge}{HTML}{C48A05}

\newcommand{\ncmlogowordbox}{\begin{tikzpicture}[x=1cm, y=1cm, line width=0.4pt]
    \def\ncmtw{0.32}\def\ncmgap{0.045}\def\ncmbar{0.16}\def\ncmtile{0.30}
    \def\ncmmidw{0.845}\def\ncmtopw{0.64}
    \pgfmathsetmacro{\ncmW}{3*\ncmtw+2*\ncmgap}
    \pgfmathsetmacro{\ncmhalf}{(\ncmmidw-\ncmgap)/2}
    \foreach \ncmL [count=\ncmi from 0] in {N,C,M} {
      \node[draw=ncmLogoEdge, fill=ncmLogoC, rounded corners=0.8pt, inner sep=0pt,
            minimum width=\ncmtw cm, minimum height=\ncmtile cm,
            font=\sffamily\bfseries\fontsize{8}{8}\selectfont, text=cNCMtxt]
        at ({\ncmi*(\ncmtw+\ncmgap)+\ncmtw/2},{\ncmtile/2}) {\ncmL};
    }
    \foreach \ncmi in {0,1} {
      \filldraw[fill=ncmLogoB, draw=ncmLogoEdge, rounded corners=0.8pt]
        ({(\ncmW-\ncmmidw)/2+\ncmi*(\ncmhalf+\ncmgap)},{\ncmtile+\ncmgap})
        rectangle ++(\ncmhalf,\ncmbar);
    }
    \filldraw[fill=ncmLogoA, draw=ncmLogoEdge, rounded corners=0.8pt]
      ({(\ncmW-\ncmtopw)/2},{\ncmtile+2*\ncmgap+\ncmbar}) rectangle ++(\ncmtopw,\ncmbar);
  \end{tikzpicture}}

\newcommand{\ncmlogo}[1][0.70cm]{\resizebox{!}{#1}{\ncmlogowordbox}}

\newcommand{\name}{\textbf{NCM}\xspace}

\tikzset{
  treenode/.style={
    align=center,
    inner sep=0pt,
    text centered,
    font=\sffamily
  },
  arn_n/.style={
    treenode,
    circle,
    black,
    font=\sffamily\bfseries,
    draw=black,
    fill=white,
    text width=1.5em
  },
  arn_r/.style={
    treenode,
    circle,
    black,
    font=\sffamily\bfseries,
    draw=black,
    fill=white,
    text width=1.0em
  },
  arn_x/.style={
    treenode,
    rectangle,
    draw=black,
    minimum width=0.5em,
    minimum height=0.5em
  }
}

\usepackage[
  colorlinks=true,
  linkcolor=linkblue,
  citecolor=linkblue,
  urlcolor=linkpink,
  filecolor=linkblue,
  bookmarks=true,
  bookmarksopen=true
]{hyperref}

\usepackage[
  nameinlink,
  capitalize,
  noabbrev
]{cleveref}

\newif\ifshownotes
\shownotestrue

\usepackage[parfill]{parskip}

\newcommand*{\ShowNotes}{}

\ifdefined\ShowNotes
  \newcommand{\colornote}[3]{{\color{#1}\bf{#2 #3}\normalfont}}
\else
  \newcommand{\colornote}[3]{}
\fi

\definecolor{darkred}{rgb}{0.7,0.1,0.1}
\definecolor{darkgreen}{rgb}{0.1,0.5,0.1}
\definecolor{cyan}{rgb}{0.7,0.0,0.7}
\definecolor{dblue}{rgb}{0.2,0.2,0.8}
\definecolor{maroon}{rgb}{0.76,.13,.28}
\definecolor{burntorange}{rgb}{0.81,.33,0}
\definecolor{royalpurple}{rgb}{0.47,.31,0.66}

\ifdefined\ShowNotes
  
\else
  
\fi

\newif\ifarxiv

\newcommand{\blfootnote}[1]{\begingroup
  \renewcommand{\thefootnote}{}\footnotetext{#1}\addtocounter{footnote}{-1}\endgroup
}

\title{\vspace{-0.5cm}\fontsize{19}{23}\selectfont\raisebox{-0.17\height}{\ncmlogo}\hspace{0.45em}\textbf{N}ext-\textbf{C}oncept \textbf{M}odeling: Fast, Parallel Coarse-to-Fine Text Generation
}

\author[1]{Hermann Kumbong\textsuperscript{*}}
\author[1]{Ishan S.\ Khare\textsuperscript{*}}
\author[1]{\protect\\ Iddah Mlauzi}
\author[2]{Andrew Woen}
\author[1]{Charles Ding}
\author[1]{E.\ Kelly Buchanan}
\author[3]{\protect\\ Simran Arora\textsuperscript{\textdagger}}
\author[1]{Azalia Mirhoseini\textsuperscript{\textdagger}}
\author[1]{Christopher R{\'e}\textsuperscript{\textdagger}}

\affil[1]{Department of Computer Science, Stanford University}
\affil[2]{Department of Electrical Engineering, Stanford University}
\affil[3]{Together AI}

\date{}

\begin{document}

\maketitle

\blfootnote{\mbox{\textsuperscript{*}Equal contribution.
    \qquad
    \textsuperscript{\textdagger}Equal senior authorship.}}
\blfootnote{Correspondence to {\hypersetup{urlcolor=black}\href{mailto:kumboh@stanford.edu,iskhare@stanford.edu}{\texttt{\{kumboh,\,iskhare\}@stanford.edu}}}.}

\vspace{-0.89cm}
{\centering\small\href{https://github.com/HazyResearch/ncm}{\textcolor{black}{\faGithub}}\enspace\url{https://github.com/HazyResearch/ncm}\par}
\vspace{4pt}

\begin{abstract}
  
\noindent Diffusion language models (DLMs) offer parallel sampling for faster generation than autoregressive models, an increasingly important advantage as inference becomes a dominant computational workload. However, DLMs degrade quickly with fewer sampling steps, process the full sequence at every step, and require many passes over the dataset to reach good quality. We introduce \textbf{Next-Concept Modeling (\name)}, a two-stage, coarse-to-fine framework: a multiscale residual VQ-VAE encodes text into a hierarchy of discrete codes, and a transformer generates one level at a time, predicting all codes within a level in parallel. At matched parameter scales, \name\ achieves better generative perplexity than the strongest DLM baseline on LM1B while running $21\times$ faster. On OpenWebText, \name\ is competitive with the strongest DLM while running $10\times$ faster; on XSum, it matches the strongest DLM's ROUGE-L while being $31\times$ faster. On the reasoning task of Sudoku, \name\ outperforms autoregressive and diffusion baselines across all difficulty levels, using fewer function evaluations.

\end{abstract}

\def\effsubshifta{15.18bp}
\def\effsubshiftb{9.30bp}
\def\effsubshiftc{17.41bp}

\newsavebox{\effpanel}\newlength{\effcapw}\newlength{\effextra}\newlength{\effgap}
\newcommand{\effsubbox}[4]{\sbox{\effpanel}{\includegraphics{#2}}\settowidth{\effcapw}{(a)\ #3}\setlength{\effextra}{\dimexpr(\effcapw+2pt-\wd\effpanel+2\dimexpr#1\relax)/2\relax}\ifdim\effextra<0pt \setlength{\effextra}{0pt}\fi
  \edef\effmargin{\noexpand\captionsetup[sub]{margin={\the\dimexpr2\dimexpr#1\relax-\effextra\relax,0pt}}}{\effmargin\subcaptionbox{#3\label{#4}}[\dimexpr\wd\effpanel+\effextra\relax][l]{\usebox{\effpanel}}}}

{\setlength{\intextsep}{8pt plus 2pt minus 2pt}
\begin{figure}[H]
\centering
\settowidth{\effgap}{\includegraphics{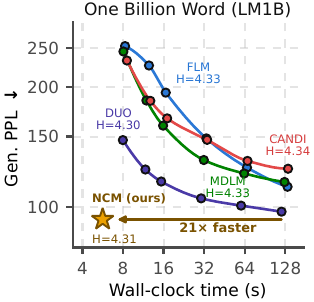}\includegraphics{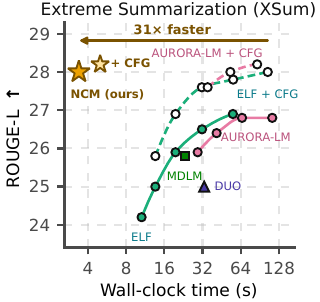}\includegraphics{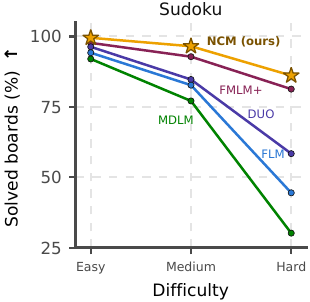}}\setlength{\effgap}{\dimexpr(\linewidth-\effgap)/2-0.5pt\relax}\effsubbox{\effsubshifta}{figs/efficiency_latency_a.pdf}{Unconditional generation}{fig:eff-lm1b}\hspace{\dimexpr\effgap-\effextra\relax}\effsubbox{\effsubshiftb}{figs/efficiency_latency_b.pdf}{Conditional generation}{fig:eff-xsum}\hspace{\dimexpr\effgap-\effextra\relax}\effsubbox{\effsubshiftc}{figs/efficiency_latency_c.pdf}{Reasoning}{fig:eff-sudoku}\hspace{-\effextra}
\caption{\textbf{Generation efficiency and reasoning quality.} \textbf{(a,b)} \name\ achieves better or on-par quality with $21\times$ and $31\times$ lower wall-clock time on LM1B and XSum, respectively (Section~\ref{subsec:efficiency}). $H$ denotes mean unigram entropy. \textbf{(c)} \name\ maintains the highest Sudoku accuracy as difficulty increases, using fewer function evaluations (NFE) than the diffusion baselines.}
\label{fig:efficiency-latency}
\end{figure}
}

\section{Introduction}
\label{sec:intro}

Diffusion language models (DLMs) are the leading non-autoregressive language
modeling paradigm \citep{austin2021d3pm,lou2024sedd,sahoo2024mdlm}. Their main
appeal is their ability to produce many tokens at a time through iterative
denoising, which can make generation substantially faster than in autoregressive (AR) models that generate one token at a time. This is becoming increasingly important, as language model inference is one of the dominant computational workloads \citep{kwon2023efficient,pope2022efficiently}. DLMs are moving from research to practice: open models
such as LLaDA \citep{nie2025llada} scale to billions of parameters, commercial
systems such as Mercury \citep{labs2025mercury} and DiffusionGemma
\citep{deepmind2025geminidiffusion} serve diffusion models in production, and
diffusion drafters like DFlash~\citep{chen2026dflash} accelerate AR models through
speculative decoding.

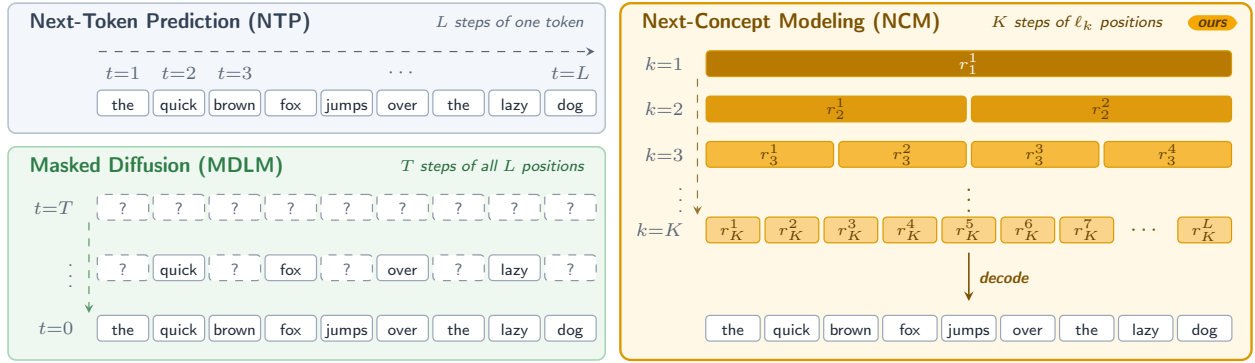
\begin{figure}[t]
  \centering
  \begingroup
\definecolor{tzInk}{HTML}{1F2937}
\definecolor{tzGrey}{HTML}{5B6778}
\definecolor{tzTokEdge}{HTML}{9AA7B8}
\definecolor{tzPanel}{HTML}{F3F6FA}
\definecolor{tzPanelEdge}{HTML}{B8C2D0}
\definecolor{tzNtpTitle}{HTML}{334155}
\definecolor{tzMdlm}{HTML}{2E7D46}
\definecolor{tzMdlmBg}{HTML}{EEF7F0}
\definecolor{tzMdlmEdge}{HTML}{A9D5B5}
\definecolor{tzNcm}{HTML}{7A5200}
\definecolor{tzNcmEdge}{HTML}{D99A00}
\definecolor{tzNcmBg}{HTML}{FFF8E6}
\definecolor{tzNcmBadge}{HTML}{F5A800}
\colorlet{tzCodeA}{cNCMlevel1}
\colorlet{tzCodeB}{cNCMlevel2}
\colorlet{tzCodeC}{cNCMlevel3}
\colorlet{tzCodeD}{cNCMlevel4}
\begin{tikzpicture}[
    x=1cm, y=1cm, font=\sffamily,
    panel/.style={draw=tzPanelEdge, fill=tzPanel, rounded corners=3pt, line width=0.6pt},
    panelG/.style={draw=tzMdlmEdge, fill=tzMdlmBg, rounded corners=3pt, line width=0.6pt},
    tok/.style={draw=tzTokEdge, fill=white, rounded corners=1.5pt, line width=0.45pt,
                minimum height=0.34cm, inner sep=0pt, font=\sffamily\tiny, text=tzInk,
                text height=1.55ex, text depth=0.45ex},
    mask/.style={tok, draw=tzGrey!70, dashed, fill=white, text=tzGrey},
    code/.style={draw=tzNcmEdge, rounded corners=1.5pt, line width=0.45pt,
                 minimum height=0.34cm, inner sep=0pt, font=\tiny, text=tzNcm!80!black},
    lbl/.style={font=\sffamily\fontsize{7}{8}\selectfont, text=tzGrey},
    arr/.style={-stealth, line width=0.45pt},
]
\def\words{{"the","quick","brown","fox","jumps","over","the","lazy","dog"}}
\def\lx{1.15}\def\lw{0.74}\def\lb{0.68}
\def\nx{9.2}\def\nw{0.78}\def\nb{0.72}

\draw[panel] (0,3.0) rectangle (7.9,4.72);
\node[anchor=west, font=\sffamily\scriptsize\bfseries, text=tzNtpTitle] at (0.15,4.46)
  {Next-Token Prediction (NTP)};
\node[anchor=east, font=\sffamily\tiny\itshape, text=tzGrey] at (7.75,4.46)
  {$L$ steps of one token};
\foreach \i in {0,...,8} {
  \pgfmathsetmacro{\w}{\words[\i]}
  \node[tok, minimum width=\lb cm] (ntp\i) at ({\lx+\i*\lw+\lw/2},3.4) {\w};
}
\foreach \i/\t in {0/1, 1/2, 2/3, 8/L} {
  \node[lbl] at ({\lx+\i*\lw+\lw/2},3.8) {$t{=}\t$};
}
\node[lbl] at ({\lx+5.5*\lw},3.8) {$\cdots$};
\draw[arr, tzGrey, dashed] ({\lx+0.05},4.08) -- ({\lx+9*\lw-0.05},4.08);

\draw[panelG] (0,0.0) rectangle (7.9,2.82);
\node[anchor=west, font=\sffamily\scriptsize\bfseries, text=tzMdlm] at (0.15,2.56)
  {Masked Diffusion (MDLM)};
\node[anchor=east, font=\sffamily\tiny\itshape, text=tzMdlm!85!black] at (7.75,2.56)
  {$T$ steps of all $L$ positions};
\def\tzmid{{0,1,0,1,0,1,0,1,0}}
\foreach \i in {0,...,8} {
  \pgfmathsetmacro{\w}{\words[\i]}
  \node[mask, minimum width=\lb cm] at ({\lx+\i*\lw+\lw/2},2.02) {?};
  \pgfmathtruncatemacro{\m}{\tzmid[\i]}
  \ifnum\m=1
    \node[tok, minimum width=\lb cm] at ({\lx+\i*\lw+\lw/2},1.22) {\w};
  \else
    \node[mask, minimum width=\lb cm] at ({\lx+\i*\lw+\lw/2},1.22) {?};
  \fi
  \node[tok, minimum width=\lb cm] at ({\lx+\i*\lw+\lw/2},0.42) {\w};
}
\node[lbl, anchor=east] at (1.0,2.02) {$t{=}T$};
\node[lbl, anchor=east] at (1.0,1.22) {$\vdots$};
\node[lbl, anchor=east] at (1.0,0.42) {$t{=}0$};
\draw[arr, tzMdlm!80, dashed] (1.07,1.82) -- (1.07,0.62);

\draw[draw=tzNcmEdge, fill=tzNcmBg, rounded corners=4pt, line width=0.9pt] (8.1,0.0) rectangle (16.45,4.72);
\node[anchor=west, font=\sffamily\scriptsize\bfseries, text=tzNcm] at (8.25,4.46)
  {Next-Concept Modeling (NCM)};
\node[fill=tzNcmBadge, text=tzNcm!70!black, rounded corners=4pt, font=\sffamily\tiny\bfseries\itshape,
      inner xsep=4pt, inner ysep=1.5pt] at (15.93,4.46) {ours};
\node[anchor=east, font=\sffamily\tiny\itshape, text=tzNcm!80!black] at (15.4,4.46)
  {$K$ steps of $\ell_k$ positions};
\newcommand{\tzlevel}[5]{\foreach \cc [count=\j from 0] in {#5} {
    \pgfmathsetmacro{\tzspan}{9/#2}
    \node[code, fill=#3, minimum width={(\tzspan*\nw-0.07)*1cm}]
      at ({\nx+(\j+0.5)*\tzspan*\nw},#1) {$r_{#4}^{\cc}$};
  }}
\tzlevel{3.92}{1}{tzCodeA, text=white}{1}{1}
\tzlevel{3.32}{2}{tzCodeB}{2}{1,2}
\tzlevel{2.72}{4}{tzCodeC}{3}{1,2,3,4}
\tzlevel{1.72}{9}{tzCodeD}{K}{1,2,3,4,5,6,7}
\node[font=\scriptsize, text=tzNcm] at ({\nx+7.5*\nw},1.72) {$\cdots$};
\node[code, fill=tzCodeD, minimum width={(\nw-0.07)*1cm}] at ({\nx+8.5*\nw},1.72) {$r_{K}^{L}$};
\node[lbl, anchor=east] at (9.07,3.92) {$k{=}1$};
\node[lbl, anchor=east] at (9.07,3.32) {$k{=}2$};
\node[lbl, anchor=east] at (9.07,2.72) {$k{=}3$};
\node[lbl, anchor=east] at (9.07,2.22) {$\vdots$};
\node[lbl, anchor=east] at (9.07,1.72) {$k{=}K$};
\node[font=\scriptsize, text=tzNcm] at ({\nx+4.5*\nw},2.22) {$\vdots$};
\draw[arr, tzNcm!80, dashed] (9.12,3.72) -- (9.12,1.92);
\draw[arr, tzNcm, line width=0.6pt] ({\nx+4.5*\nw},1.42) -- ({\nx+4.5*\nw},0.78)
  node[midway, right, font=\sffamily\tiny\itshape\bfseries, text=tzNcm] {decode};
\foreach \i in {0,...,8} {
  \pgfmathsetmacro{\w}{\words[\i]}
  \node[tok, minimum width=\nb cm] at ({\nx+\i*\nw+\nw/2},0.42) {\w};
}
\end{tikzpicture}
\endgroup
  \caption{\textbf{Comparing text generation paradigms.} Next-token prediction
    generates one token per step, and masked diffusion (discrete diffusion) refines all $L$ positions
    at every step. \name instead generates discrete codes $r_k^i$, the $i$-th code of level $k$,
    from coarse to fine in $K \ll L$ steps: each step predicts the $\ell_k$ codes
    of one level in parallel, conditioned on the reconstruction $\hat z^{(k-1)}$
    of the earlier levels, and after the final level, the cumulative reconstruction is decoded into tokens.}
  \label{fig:teaser}
\end{figure}

The promise of DLMs is parallel sampling: generating a sequence in far
fewer sequential steps than it has tokens. In practice however:
\begin{itemize}[leftmargin=*, topsep=4pt, itemsep=3pt, parsep=0pt]
\item \textbf{(i) Many sampling steps.} Diffusion models predict tokens in parallel
but still perform successive denoising passes. Reducing these passes can substantially
degrade generation quality (Figure~\ref{fig:gen-ppl}), limiting the speed gains from parallelism.
\item \textbf{(ii) Expensive sampling steps.} Processing the full sequence at every
step limits speedups, and bidirectional attention prevents direct reuse of AR-style
KV caching~\citep{wu2025fastdllm,ma2025dkvcache}.
\item \textbf{(iii) Large training budgets.} DLM training often involves many passes
over the same data. For example, strong diffusion baselines such as
DUO and MDLM are trained on OpenWebText for $1$M steps, processing over $500$B
tokens \citep{sahoo2025duo,sahoo2024mdlm,hu2026elf}.
\end{itemize}

Prior work reduces the number of sampling steps mainly through few-step samplers and
distillation, either by distilling a many-step teacher into a few-step student
\citep{sahoo2025duo,deschenaux2025sdtt} or by learning a direct map between noise
levels \citep{lee2026flm}.
These approaches come at the cost of additional training stages on top of an already
expensive model. Few-step sampling also comes at a price: distilled models tend to be less diverse
than the models they are distilled from. On standard benchmarks such as OpenWebText
\citep{Gokaslan2019OpenWeb} and LM1B \citep{chelba2013one}, they reach a lower
generative perplexity at the expense of sample diversity, as measured by mean
per-sample unigram entropy \citep{lee2026flm}. Moreover, these approaches leave the per-step cost unchanged.
Block-autoregressive diffusion enables KV caching at the cost of parallelism:
tokens within each block can be denoised in parallel, but blocks must be generated
sequentially, with each waiting for the previous block to finish
\citep{arriola2025block,wu2026fastdllmv2}.

We introduce \textbf{Next-Concept Modeling (\name)}, a two-stage language modeling
framework. A multiscale residual VQ-VAE \citep{vandenoord2017vqvae} encodes text into
$K$ levels of discrete codes, from short coarse representations to longer fine ones.
A transformer generates these levels sequentially, predicting all codes within each
level in parallel, following next-scale prediction for images
\citep{tian2024var,kumbong2025hmar}. \name\ achieves \textbf{(i)} high generation
quality and diversity in $K \ll L$ steps without distillation (Figure~\ref{fig:gen-ppl});
\textbf{(ii)} lower sampling cost by processing short code sequences at early levels
(Figure~\ref{fig:efficiency-latency});
and \textbf{(iii)} strong performance with fewer training passes over the data
(Figure~\ref{fig:lm1b_tokens}).

We evaluate \name\ models at approximately $100$M parameters against discrete-diffusion,
flow-based, and hybrid baselines, following standard evaluation protocols
\citep{sahoo2024mdlm,sahoo2025duo}. Across unconditional generation, conditional
generation (translation and summarization), and reasoning tasks (Sudoku), we find:
\begin{itemize}
\item \textbf{(i) High quality in few steps.} Across three unconditional generation
datasets (TinyStories, LM1B, and OpenWebText), \name matches or outperforms strong
diffusion baselines with approximately $4$--$11\times$ fewer function evaluations
(NFE), while maintaining sample entropy close to the data (Figure~\ref{fig:gen-ppl}).
On conditional generation (XSum summarization and WMT14 translation), it uses
approximately $6$--$16\times$ fewer NFE than unguided baselines, achieving the best
summarization scores and outperforming all but one non-AR translation baseline
(Table~\ref{tab:seq2seq}). On Sudoku, \name outperforms AR and diffusion baselines across
difficulty levels while using fewer NFE than any of them (Table~\ref{tab:reasoning}).
\item \textbf{(ii) Fast inference and effective test-time scaling.} Fewer sampling
steps and lower per-step cost from shorter coarse levels yield $10$--$31\times$
faster generation at comparable or better quality (Figures~\ref{fig:efficiency-latency}
and~\ref{fig:gen-ppl}). Using the saved time to generate more candidates and select
among them with MBR decoding yields about $7\%$ higher ROUGE-L than the strongest
summarization baseline within the same sampling-time budget
(Figure~\ref{fig:efficiency-group}).
\item \textbf{(iii) Lower training budgets.} \name achieves competitive generation
quality with approximately $4$--$5\times$ fewer training tokens than strong diffusion
baselines on LM1B and OpenWebText, including both tokenizer and generator training
(Figure~\ref{fig:lm1b_tokens}; Appendix~\ref{app:architecture}).
\end{itemize}
We also show preliminary scaling results: increasing \name's model size improves
generation quality on LM1B at a fixed training-token budget
(Figure~\ref{fig:scaling}).
Qualitative probes suggest that coarse levels capture global structure while finer
levels add detail
(Appendix~\ref{app:hierarchy}, Figures~\ref{fig:story-continuation}--\ref{fig:story-perturbation}).

\begin{figure}[!t]
\centering
\begingroup
\definecolor{trInk}{HTML}{1F2937}
\definecolor{trMuted}{HTML}{5B6778}
\definecolor{trLine}{HTML}{D6DEE8}
\definecolor{trArrow}{HTML}{8C98A8}
\definecolor{trGood}{HTML}{0CA30C}
\colorlet{trHit}{trGood!24}
\newcommand{\trhit}[1]{{\setlength{\fboxsep}{0.8pt}\colorbox{trHit}{\vphantom{Ag}#1}}}
\begin{tcolorbox}[enhanced,arc=0pt,boxrule=0.8pt,
  colback=white,colframe=black,
  left=2.5mm,right=2.5mm,top=1.8mm,bottom=1.8mm,
  before skip=0pt,after skip=7pt]
\scriptsize\linespread{1.06}\selectfont
{\footnotesize\textbf{Translation}}\hfill{\scriptsize\textbf{BLEU:} 77.8}\par
\vspace{1pt}{\color{black}\hrule height 0.8pt}\vspace{4pt}
\resizebox{\linewidth}{!}{\begin{tabular}{@{}l@{\hspace{6pt}}l@{}}
\textbf{German:} & Es besteht allerdings die Sorge, dass kleine und mittlere Unternehmen
  angreifbar gegenüber Hacking und Überwachung bleiben. \\[2pt]
\textbf{Reference:} & However, there is concern that small and medium-sized companies remain
  vulnerable to hacking and surveillance. \\
\end{tabular}}
\end{tcolorbox}
\begin{tikzpicture}[x=1cm,y=1cm,
    body/.style={anchor=north west,text=trInk,inner sep=0pt,
      font=\sffamily\fontsize{7}{8.9}\selectfont}]
\pgfmathsetmacro{\trG}{0.34}
\pgfmathsetmacro{\trW}{(\linewidth/1cm-0.06-4*\trG)/5}
\pgfmathsetmacro{\trH}{3.15}
\newcommand{\trbox}[5]{\pgfmathsetmacro{\x}{0.03+#1*(\trW+\trG)}
  \fill[#2] (\x,0) rectangle ++(\trW,-0.13);
  \draw[#4] (\x,0) rectangle ++(\trW,-\trH);
  \node[anchor=base west,inner sep=0pt,text=trInk,
        font=\sffamily\fontsize{8}{10}\selectfont\bfseries] at ({\x+0.16},-0.47) {Step #3};
  \draw[trLine,line width=0.5pt] ({\x+0.16},-0.62) -- ({\x+\trW-0.16},-0.62);
  \node[body] at ({\x+0.16},-0.78) {\begin{minipage}{\dimexpr\trW cm-0.32cm\relax}\raggedright
    \hyphenpenalty=10000\exhyphenpenalty=10000 #5\end{minipage}};
  \ifnum#1>0
    \draw[-stealth,trArrow,line width=0.55pt] ({\x-\trG+0.05},{-\trH/2}) -- ({\x-0.05},{-\trH/2});
  \fi}
\node[anchor=base west,inner sep=0pt,text=trMuted,
      font=\sffamily\fontsize{7}{8.5}\selectfont] at (0,0.22)
  {Decoded output after each sampling step, coarse $\to$ fine};
\trbox{0}{cNCMlevel1}{1}{black,line width=0.8pt}{\trhit{However,} it \trhit{is} a important \trhit{to} the the the \trhit{and} the.}
\trbox{1}{cNCMlevel2}{3}{black,line width=0.8pt}{\trhit{However,} it \trhit{is} a \trhit{concern} for \trhit{small} small small
  \trhit{medium-sized} and-sized-sized \trhit{and} \trhit{and} and...}
\trbox{2}{cNCMlevel3}{6}{black,line width=0.8pt}{\trhit{However,} \trhit{there} \trhit{is} a \trhit{concern} \trhit{that} medium
  \trhit{medium-sized} enterprises will \trhit{remain} remain remain \trhit{surveillance}
  surveillance surveillance capabilities}
\trbox{3}{cNCMlevel3!50!cNCMlevel4}{8}{black,line width=0.8pt}{\trhit{However,} \trhit{there} \trhit{is} \trhit{that} \trhit{concern} \trhit{and}
  \trhit{and} \trhit{medium-sized} enterprises will \trhit{remain} \trhit{vulnerable}
  \trhit{to} \trhit{hacking} and \trhit{surveillance.}}
\trbox{4}{cNCMlevel4}{11}{cNCMlevel1,line width=1.4pt}{\trhit{However,} \trhit{there} \trhit{is} \trhit{concern} \trhit{that} \trhit{small}
  \trhit{and} \trhit{medium-sized} enterprises will \trhit{remain} \trhit{vulnerable}
  \trhit{to} \trhit{hacking} \trhit{and} \trhit{surveillance.}}
\end{tikzpicture}
\endgroup
\caption{\textbf{NCM generation trajectory.} Output decoded after selected sampling steps of one
German--English translation (WMT14), shown verbatim; the last step (gold frame) is the final
output. Words matching the reference are shaded green. Coarse levels fix the high-level idea; finer levels fill in and correct
the content. Appendix~\ref{app:translation-trajectory} shows all sampling steps with their scores.}
\label{fig:generation-trajectory}
\end{figure}
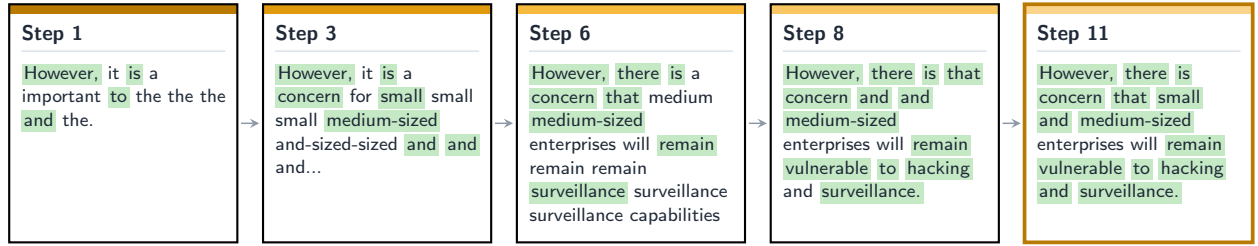

\section{Background \& Related Work}
\label{sec:related}

\paragraph{Diffusion \& Flow-Based Models for Text Generation.}
Text generation using Diffusion \& Flow-Based Models is an iterative refinement
process. Discrete methods operate on token sequences
\citep{austin2021d3pm,lou2024sedd,sahoo2024mdlm}; continuous methods evolve
representations of tokens in an embedding space
\citep{hu2026elf,dieleman2022cdcd,pynadath2025candi,liang2026aurora}.
Distillation/flow-map techniques enable reduction of the number of sampling steps
\citep{sahoo2025duo,deschenaux2025sdtt,lee2026flm}; however, full-sequence
formulations still require processing a length-$L$ representation in every step.
Block-Autoregressive Diffusion allows KV caching through sequential block
generation: tokens within a block are generated in parallel; however, each block
must be generated after the previous blocks
\citep{arriola2025block,wu2026fastdllmv2}. \name modifies both the representation
and the generation schedule. \name generates a learned hierarchy of discrete
codes from coarse to fine, and predicts all codes within each level in parallel.
Generation therefore follows a short sequence of levels, with fewer positions
processed at the early, coarse levels.

\paragraph{Discrete latent variables and next scale prediction.}
A number of image generative models use discrete tokenizers
\citep{vandenoord2017vqvae,razavi2019vqvae2,esser2021taming}. Residual quantization
corrects for errors made at previous steps and uses those errors to create a new
codebook \citep{lee2022rqvae} which has been adapted to neural audio codecs
\citep{zeghidour2021soundstream,defossez2022encodec}. Next-scale
prediction generates such hierarchies coarse to fine, one scale at a time in parallel, for
fast, high-quality images \citep{tian2024var,kumbong2025hmar}. \name extends residual
quantization and next-scale prediction to text through a hierarchy of discrete
codes that determines the generation order. While latent diffusion models generate
continuous representations of text \citep{lovelace2023ld4lg,zhang2023planner},
\name produces successive discrete levels. Adapting this approach to text requires
learning useful coarse representations without a natural counterpart to image
downsampling. Accurate reconstruction is also important: a single token error can
alter meaning, prevent code from running, or invalidate a Sudoku solution.

\paragraph{Hierarchical and concept-level text generation.}
Several approaches organize language modeling around larger units and learned
abstractions. MEGABYTE~\citep{yu2023megabytepredictingmillionbytesequences}
combines global modeling of byte patches with local byte prediction, while
H-Net~\citep{hwang2025dynamicchunkingendtoendhierarchical} learns dynamic chunks
within a multistage hierarchy. Large Concept Models~\citep{barrault2024large}
predict sentence representations. ConceptLM~\citep{liu2026next} predicts discrete
concepts to guide subsequent token generation, and Dynamic Large Concept
Models~\citep{qu2025dynamic} learn variable-length concepts for computation in a
compressed representation space. HDLM~\citep{zhou2025next} progressively refines
tokens through a hierarchy of semantic vocabularies. \name represents an entire
sequence through discrete levels of increasing resolution and generates these
levels in order using a shared transformer. Predicting all codes within each
level in parallel gives a short generation path, while the shorter coarse levels
reduce computation during early steps.

\section{Next-Concept Modeling}
\label{sec:method}

\name generates text through a learned coarse-to-fine hierarchy of discrete codes.
We first train a multiscale residual VQ tokenizer to represent text at progressively
finer levels of detail, then freeze it and train a transformer to generate the hierarchy.
At each level, the transformer predicts all codes in parallel, conditioned on the
cumulative reconstruction of earlier levels. The tokenizer's decoder maps the
completed reconstruction back to text. Figure~\ref{fig:generation-trajectory}
previews this process through a German--English translation, showing how the
output develops across levels.
The term \emph{concept} here refers to the information carried by the coarse
levels of the hierarchy.

\begin{figure}[!t]
    \centering
    \begingroup
\definecolor{arInk}{HTML}{1F2937}
\definecolor{arGrey}{HTML}{5B6778}
\definecolor{arPanelEdge}{HTML}{B8C2D0}
\definecolor{arTokEdge}{HTML}{9AA7B8}
\definecolor{arArrow}{HTML}{4A5568}
\definecolor{arTrap}{HTML}{E9EDF2}
\definecolor{arTrapEdge}{HTML}{8C98A8}
\definecolor{arNcm}{HTML}{7A5200}
\definecolor{arNcmEdge}{HTML}{D99A00}
\definecolor{arNcmBg}{HTML}{FFF8E6}
\definecolor{arCellEdge}{HTML}{B07A0A}
\definecolor{cgA}{HTML}{E09A06}
\definecolor{cgB}{HTML}{F5B42A}
\definecolor{cgC}{HTML}{F9D27A}
\definecolor{cgD}{HTML}{FDEBC0}
\definecolor{ltA}{HTML}{93A3B8}
\definecolor{ltB}{HTML}{B4C1D1}
\definecolor{ltC}{HTML}{D3DCE6}
\definecolor{ltD}{HTML}{E9EEF3}
\definecolor{arLatBg}{HTML}{F3F6FA}
\definecolor{ctxA}{HTML}{7FA6D6}
\definecolor{ctxB}{HTML}{A9C4E6}
\definecolor{arCtxBg}{HTML}{EEF4FB}
\definecolor{arCtxEdge}{HTML}{6F93C2}
\definecolor{arTrans}{HTML}{0E3A6B}
\definecolor{arLoss}{HTML}{46505E}
\definecolor{arMask}{HTML}{F1F2F4}
\begin{tikzpicture}[
    x=1cm, y=1cm, font=\sffamily,
    panel/.style={draw=arPanelEdge, fill=white, rounded corners=3pt, line width=0.6pt},
    title/.style={anchor=west, font=\sffamily\scriptsize\bfseries, text=arInk},
    tok/.style={draw=arTokEdge, fill=white, rounded corners=1.5pt, line width=0.45pt,
                minimum height=0.34cm, minimum width=1.1cm, inner sep=0pt,
                font=\sffamily\tiny, text=arInk},
    cell/.style={draw=arCellEdge, line width=0.3pt, rounded corners=0.6pt,
                 minimum size=0.26cm, inner sep=0pt},
    latcell/.style={cell, draw=arTrapEdge},
    ctxcell/.style={cell, draw=arCtxEdge},
    grp/.style={draw=arNcmEdge!75, fill=arNcmBg, rounded corners=1.8pt, line width=0.45pt},
    latgrp/.style={draw=arTokEdge, fill=arLatBg, rounded corners=1.8pt, line width=0.45pt},
    ctxgrp/.style={draw=arCtxEdge, fill=arCtxBg, rounded corners=1.8pt, line width=0.45pt},
    arr/.style={-stealth, line width=0.5pt, arArrow},
    rl/.style={font=\scriptsize, text=arNcm},
    lbl/.style={font=\sffamily\fontsize{6.5}{7.5}\selectfont, text=arGrey},
    hd/.style={font=\sffamily\fontsize{6.5}{7.5}\selectfont\bfseries, text=arInk},
]
\def\ardots{D}
\newcommand{\arblock}[7]{\pgfmathsetmacro{\arbw}{#4*0.3-0.04}\draw[#6] ({#2-0.05},{#3-0.18}) rectangle ({#2+\arbw+0.05},{#3+0.18});
  \coordinate (#1) at ({#2+\arbw/2},#3);
  \coordinate (#1-n) at ({#2+\arbw/2},{#3+0.18});
  \coordinate (#1-s) at ({#2+\arbw/2},{#3-0.18});
  \foreach \aritem [count=\arpos from 0] in {#5} {
    \ifx\aritem\ardots
      \node[font=\tiny, text=arGrey] at ({#2+\arpos*0.3+0.13},#3) {$\cdots$};
    \else
      \node[#7, fill=\aritem] at ({#2+\arpos*0.3+0.13},#3) {};
    \fi
  }}

\draw[panel] (0,0) rectangle (7.95,6.6);
\node[title] at (0.15,6.33) {\textcolor{arGrey}{(a)}\ Stage 1: Learn a Multiscale Tokenizer};

\node[draw=arLoss!60, fill=arLoss!7, rounded corners=2pt, line width=0.5pt,
      font=\sffamily\fontsize{6.5}{7.5}\selectfont\bfseries, text=arLoss,
      inner xsep=5pt, inner ysep=2.5pt] (loss) at (3.975,5.72) {Reconstruction loss};
\draw[stealth-, arLoss, line width=0.6pt] (loss.west) -| (0.8,5.12);
\draw[stealth-, arLoss, line width=0.6pt] (loss.east) -| (7.12,5.12);

\node[hd] at (0.8,4.93) {Input $x_{1:L}$};
\node[hd] at (7.12,4.93) {Output $\hat x_{1:L}$};
\foreach \w/\y in {Madison/4.5, Square/4.05, Garden/3.6, was/3.15, the/2.7, place/2.25,
                   to/1.8, York/1.0, Knicks./0.55} {
  \node[tok] at (0.8,\y) {\w};
  \node[tok] at (7.12,\y) {\w};
}
\node[lbl] at (0.8,1.43) {$\vdots$};
\node[lbl] at (7.12,1.43) {$\vdots$};

\filldraw[fill=arTrap, draw=arTrapEdge, line width=0.5pt, rounded corners=0.8pt]
  (1.58,1.6) -- (1.58,3.45) -- (2.18,2.95) -- (2.18,2.1) -- cycle;
\node[hd] at (1.88,3.72) {Encoder};
\filldraw[fill=arTrap, draw=arTrapEdge, line width=0.5pt, rounded corners=0.8pt]
  (5.72,2.1) -- (5.72,2.95) -- (6.32,3.45) -- (6.32,1.6) -- cycle;
\node[hd] at (6.02,3.72) {Decoder};
\draw[arr] (1.37,2.525) -- (1.56,2.525);
\draw[arr] (2.2,2.525) -- (2.38,2.525);
\draw[arr] (5.52,2.525) -- (5.70,2.525);
\draw[arr] (6.34,2.525) -- (6.55,2.525);

\draw[dashed, draw=arNcmEdge, fill=arNcmBg!45, rounded corners=2.5pt, line width=0.55pt]
  (2.4,0.98) rectangle (5.5,4.2);
\node[font=\sffamily\fontsize{6.5}{7.5}\selectfont\bfseries, text=arNcm] at (3.95,3.97)
  {Multiscale residual codes};
\node[lbl] at (3.95,3.72) {coarse $\rightarrow$ fine};
\node[rl, anchor=east] at (2.95,3.25) {$r_1$};
\node[rl, anchor=east] at (2.95,2.75) {$r_2$};
\node[rl, anchor=east] at (2.95,2.25) {$r_3$};
\node[rl, anchor=east] at (2.95,1.83) {$\vdots$};
\node[rl, anchor=east] at (2.95,1.40) {$r_K$};
\arblock{s1r1}{3.02}{3.25}{1}{cgB}{grp}{cell}
\arblock{s1r2}{3.02}{2.75}{2}{cgA,cgC}{grp}{cell}
\arblock{s1r3}{3.02}{2.25}{4}{cgD,cgB,cgA,cgC}{grp}{cell}
\arblock{s1rK}{3.02}{1.40}{8}{cgC,cgA,cgD,cgB,cgA,cgB,cgC,cgD}{grp}{cell}

\draw[panel] (8.15,0) rectangle (16.45,6.6);
\node[title] at (8.3,6.33) {\textcolor{arGrey}{(b)}\ Stage 2: Learn to Generate};

\draw[dashed, draw=arTokEdge, rounded corners=2pt, line width=0.5pt] (8.84,4.1) rectangle (13.96,5.98);
\arblock{l1T}{8.95}{5.38}{1}{cgB}{grp}{cell}
\arblock{l2T}{9.72}{5.38}{2}{cgA,cgC}{grp}{cell}
\arblock{l3T}{10.80}{5.38}{4}{cgD,cgB,cgA,cgC}{grp}{cell}
\node[lbl] at (12.34,5.38) {$\cdots$};
\arblock{lKT}{12.65}{5.38}{4}{cgC,cgA,D,cgD}{grp}{cell}
\arblock{l1P}{8.95}{4.46}{1}{cgB}{grp}{cell}
\arblock{l2P}{9.72}{4.46}{2}{cgA,cgC}{grp}{cell}
\arblock{l3P}{10.80}{4.46}{4}{cgD,cgC,cgA,cgC}{grp}{cell}
\node[lbl] at (12.34,4.46) {$\cdots$};
\arblock{lKP}{12.65}{4.46}{4}{cgC,cgA,D,cgB}{grp}{cell}
\foreach \b/\k in {l1/1, l2/2, l3/3, lK/K} {
  \node[rl] at (\b T-n) [above=0.5pt] {$r_\k$};
  \node[rl] at (\b P-n) [above=0.5pt] {$\hat r_\k$};
}
\draw[stealth-stealth, arLoss, line width=0.6pt] (8.72,4.46) -- (8.72,5.38);
\node[rotate=90, font=\sffamily\fontsize{6.5}{7.5}\selectfont\bfseries, text=arLoss]
  at (8.53,4.92) {Cross-entropy};

\fill[arTrans, rounded corners=3pt] (8.72,2.63) rectangle (13.89,3.31);
\node[font=\sffamily\scriptsize\bfseries, text=white] at (11.305,2.97) {Next-Concept Transformer};
\foreach \b in {l1P, l2P, l3P, lKP} {
  \draw[arr] (\b-s) ++(0,-0.93) -- ++(0,0.89);
}

\arblock{l1In}{8.95}{1.62}{1}{ltD}{latgrp}{latcell}
\arblock{l2In}{9.72}{1.62}{2}{ltB,ltC}{latgrp}{latcell}
\arblock{l3In}{10.80}{1.62}{4}{ltC,ltA,ltB,ltD}{latgrp}{latcell}
\node[lbl] at (12.34,1.62) {$\cdots$};
\arblock{lKIn}{12.65}{1.62}{4}{ltB,ltC,D,ltA}{latgrp}{latcell}
\foreach \b in {l1In, l2In, l3In, lKIn} {
  \draw[arr] (\b-n) ++(0,0.04) -- ++(0,0.75);
}
\node[font=\scriptsize, text=arInk] at (l1In-s) [below=1.5pt] {$\hat z^{(0)}$};
\node[font=\scriptsize, text=arInk] at (l2In-s) [below=1.5pt] {$\hat z^{(1)}$};
\node[font=\scriptsize, text=arInk] at (l3In-s) [below=1.5pt] {$\hat z^{(2)}$};
\node[font=\scriptsize, text=arInk] at (lKIn-s) [below=1.5pt] {$\hat z^{(K-1)}$};
\node[lbl, font=\sffamily\fontsize{6.5}{7.5}\selectfont\itshape, align=center] at (11.30,0.55)
  {cumulative reconstruction of the earlier levels,\\ resized to the $\ell_k$ positions of level $k$};

\def\arblk{{1,2,2,3,3,3,3,4,4,4,4,4,4,4,4}}
\def\ars{0.13}\def\armx{14.35}\def\army{3.80}
\node[font=\sffamily\scriptsize\bfseries\itshape, text=arInk, align=center] at (15.33,4.27)
  {Block-diagonal\\ attention};
\draw[dashed, arGrey, line width=0.45pt] (13.89,2.97) -- (\armx,2.97);
\foreach \mrow in {0,...,14} {
  \foreach \mcol in {0,...,14} {
    \pgfmathtruncatemacro{\arbr}{\arblk[\mrow]}
    \pgfmathtruncatemacro{\arbq}{\arblk[\mcol]}
    \ifnum\arbr=\arbq \edef\arfill{cNCMlevel\arbq}\else \def\arfill{arMask}\fi
    \filldraw[fill=\arfill, draw=white, line width=0.25pt]
      ({\armx+\mcol*\ars},{\army-\mrow*\ars}) rectangle ++(\ars,-\ars);
  }
}
\draw[arTokEdge, line width=0.4pt] (\armx,\army) rectangle ++({15*\ars},{-15*\ars});
\draw[arGrey, line width=0.4pt] ({\armx+0.02},{\army-15*\ars-0.07}) -- ++(0,-0.05)
  -- ({\armx+15*\ars-0.02},{\army-15*\ars-0.12}) -- ++(0,0.05);
\node[lbl] at ({\armx+7.5*\ars},{\army-15*\ars-0.26}) {levels $1,\ldots,K$};
\foreach \lcol/\ltext/\ly in {arMask/masked/0.68} {
  \filldraw[fill=\lcol, draw=arTokEdge, line width=0.3pt] (14.40,{\ly-0.06}) rectangle ++(0.12,0.12);
  \node[lbl, anchor=west] at (14.55,\ly) {\ltext};
}
\foreach \li in {1,...,4} {
  \fill[cNCMlevel\li] ({14.40+(\li-1)*0.035},0.87) rectangle ++(0.035,0.12);
}
\draw[arTokEdge, line width=0.3pt] (14.40,0.87) rectangle ++(0.14,0.12);
\node[lbl, anchor=west] at (14.55,0.93) {own level};
\end{tikzpicture}
\endgroup
    \caption{\textbf{\name's two-stage training.} \textbf{(a)}~A multiscale residual VQ-VAE learns to encode text into a coarse-to-fine hierarchy of discrete codes $r_1,\ldots,r_K$ and reconstruct the input. \textbf{(b)}~The \name\ transformer predicts all codes within level $k$ in parallel from the cumulative reconstruction $\hat z^{(k-1)}$, resized to $\ell_k$ positions. In the unconditional setting, block-diagonal attention restricts each input block to itself. During training, all $K$ levels are predicted in one forward pass using a cross-entropy loss. During generation, levels are predicted sequentially from coarse to fine.}
    \label{fig:method}
\end{figure}
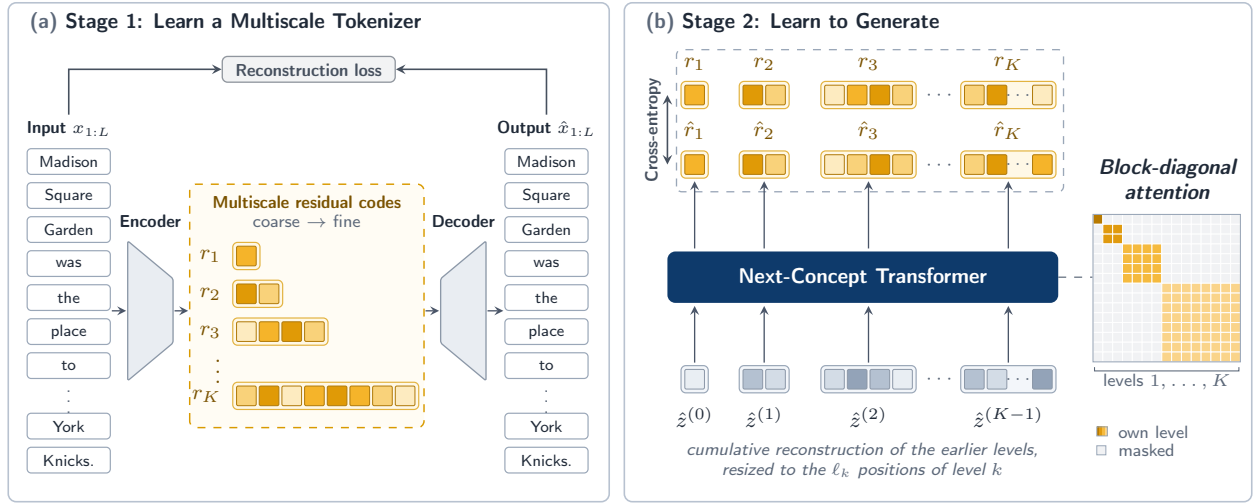

\subsection{Preliminaries}
\label{subsec:preliminaries}
Given latent vectors $z\in\mathbb{R}^{L\times d}$ and a codebook
$\mathcal V=\{e_1,\ldots,e_{|\mathcal V|}\}$, with entries
$e_j\in\mathbb{R}^d$, \emph{vector quantization} \citep{vandenoord2017vqvae}
replaces each vector $z^i$ with its nearest codebook entry:
\begin{equation}
z_q^i=\arg\min_{e\in\mathcal V}\|z^i-e\|_2^2
\end{equation}
The selected codebook vectors form the quantized latent
$z_q\in\mathbb{R}^{L\times d}$.

A single quantization step generally leaves an approximation error. \emph{Residual
quantization} \citep{zeghidour2021soundstream} represents this error using
additional codebooks, adding each quantized correction to the reconstruction.
Starting from the residual $z^{(0)}=z$
and the reconstruction $\hat z^{(0)}=0$, let $z^{(k),i}$ denote the residual
vector at position $i$ after level $k$, and $\hat z^{(k),i}$ the corresponding
cumulative reconstruction vector. Level $k$ quantizes the residual into
$L$ codes $r_k$ and updates the reconstruction and residual. Here $e_k(r_k)$
denotes lookup in level $k$'s codebook, with $e_k(r_k)^i=e_{k,r_k^i}$:
\begin{equation}
\hat z^{(k)}=\hat z^{(k-1)}+e_k(r_k),\qquad z^{(k)}=z-\hat z^{(k)}
\end{equation}
The sum $\hat z^{(K)}=\sum_{k=1}^{K}e_k(r_k)$ approximates $z$, and each partial sum
$\hat z^{(k)}$, the \emph{cumulative reconstruction}, is a partial reconstruction
using the first $k$ levels.
\emph{Multiscale residual quantization}
\citep{tian2024var} uses progressively longer code sequences, with
level $k$ containing $\ell_k$ codes. Its codebook vectors are mapped to a
length-$L$ contribution by a reconstruction operator $R_k$ before being added
to the cumulative reconstruction, allowing
shorter, coarse levels to be followed by finer levels that add detail.

\subsection{Formulation}
\label{sec:method:ncm}
Let $x\in\mathcal T^L$ be a sequence of $L$ tokens from a text vocabulary $\mathcal T$. The tokenizer maps $x$ to discrete codes at $K$ levels,
\begin{equation}
r_{1:K} = (r_1,\ldots,r_K), \qquad r_k \in \{1,\ldots,|\mathcal V_k|\}^{\ell_k},
\end{equation}
where $\mathcal V_k$ is level $k$'s codebook, $|\mathcal V_k|$ is its size,
$\ell_k$ is the number of codes at that level, and
$\ell_1 < \ell_2 < \cdots < \ell_K=L$. Each level encodes the residual left
by the earlier ones. The generator predicts the levels from coarse to fine,
conditioning on earlier levels through their cumulative reconstruction
$\hat z^{(k-1)}$:
\begin{equation}
p_\theta(r_{1:K})
= \prod_{k=1}^{K} p_\theta(r_k\mid \hat z^{(k-1)})
\end{equation}
Conditional on this reconstruction, the codes within each
level are sampled independently, allowing all positions to be generated in
one forward pass:
\begin{equation}
p_\theta(r_k\mid \hat z^{(k-1)})
=\prod_{i=1}^{\ell_k}p_\theta(r_k^i\mid \hat z^{(k-1)})
\end{equation}
After all levels are generated, the tokenizer's decoder maps their cumulative
reconstruction $\hat z^{(K)}$ to token distributions at all $L$ positions.

\subsection{Multiscale Residual Tokenization}
\label{subsec:tokenizer}
To obtain the hierarchy, we learn a multiscale residual VQ-VAE, using nested
dropout to encourage informative partial reconstructions.
An encoder $E_\phi$ maps the input sequence $x$ to $L$ latent vectors of
dimension $d$:
\begin{equation}
z = E_\phi(x) \in \mathbb{R}^{L \times d}
\end{equation}
Starting from $z^{(0)}=z$ and $\hat z^{(0)}=0$, each level quantizes the residual left by the preceding levels. At level $k$, the residual is downsampled to $\ell_k$ latent vectors, and each vector is assigned the index of its nearest entry in the codebook $\mathcal V_k$:
\begin{equation}
u_k=\mathrm{Down}_k(z^{(k-1)})\in\mathbb{R}^{\ell_k\times d},\qquad
r_k^i=\arg\min_{j\in\{1,\ldots,|\mathcal V_k|\}}
\|u_k^i-e_{k,j}\|_2^2
\end{equation}
The reconstruction operator $R_k$ maps the selected codebook vectors to a
length-$L$ contribution, which we add to the cumulative reconstruction:
\begin{equation}
\hat z^{(k)}=\hat z^{(k-1)}+
R_k(e_k(r_k)),
\qquad z^{(k)}=z-\hat z^{(k)}
\end{equation}

Here $e_k(r_k)\in\mathbb{R}^{\ell_k\times d}$ contains the selected codebook
vectors, and both $\hat z^{(k)}$ and $z^{(k)}$ have shape $L\times d$.

The VQ-VAE decoder defines token distributions
$p_\psi(x^i\mid\hat z^{(k)})$ at all $L$ positions from a cumulative
reconstruction $\hat z^{(k)}$.

\paragraph{Nested dropout.}
We use nested dropout
\citep{li2024imagefolder}, which during training retains only the first $k$ levels for a sampled
cutoff $k\in\{1,\ldots,K\}$, giving
$\hat z^{(k)}=\sum_{j=1}^{k}R_j(e_j(r_j))$.
This encourages earlier levels to capture meaningful information rather than
relying on finer levels to carry it.
During training, both the reconstruction cross-entropy
and the VQ terms are evaluated at the sampled cutoff:
\begin{equation}
\mathcal L_{\mathrm{tok}}
=-\frac{1}{L}\sum_{i=1}^{L}
\log p_\psi(x^i\mid\hat z^{(k)})
+\beta\,\mathrm{MSE}\!\left(z,\mathrm{sg}[\hat z^{(k)}]\right)
+\mathrm{MSE}\!\left(\mathrm{sg}[z],\hat z^{(k)}\right)
\end{equation}
The VQ terms combine a commitment loss and a latent quantization loss, both
measured by mean squared error (MSE).
Here $\mathrm{sg}$ denotes stop-gradient and
$\beta$ weights the commitment term.
We use a straight-through estimator to pass reconstruction gradients through
the discrete quantization step to the encoder.

\subsection{Coarse-to-Fine Generation}
\label{sec:architecture}
\label{subsec:transformer}
With the tokenizer frozen, we train the \name\ transformer on all levels in one forward pass (Algorithm~\ref{alg:ncm-training}); generation proceeds from coarse to fine in $K$ steps (Algorithm~\ref{alg:ncm-generation}).

\paragraph{Conditioning Corruption.}
\label{subsec:training}
To improve robustness to accumulating VQ code errors during generation, we train with a fraction of conditioning codes replaced by other codes from the same level's codebook, keeping the ground-truth codes intact as labels. Replacement probabilities vary by level; ablations are given in Appendix~\ref{app:corruption}. Let $\tilde z^{(k-1)}$ denote the resulting cumulative reconstruction. The training objective is
\begin{equation}
\mathcal L_{\mathrm{NCM}}
=-\sum_{k=1}^{K}w_k\sum_{i=1}^{\ell_k}
\log p_\theta(r_k^i\mid \tilde z^{(k-1)}),
\qquad
w_k=\frac{\ell_k^{-\alpha}}{\sum_{j=1}^{K}\ell_j^{1-\alpha}}
\end{equation}
With $\alpha=0$, all code positions have equal weight, and with $\alpha=1$, each level has equal weight.

\begingroup
\setlength{\parindent}{0pt}
\setlength{\parfillskip}{0pt}
\noindent\makebox[\textwidth][s]{\begin{minipage}[t]{0.44\textwidth}
\begin{algorithm}[H]
\captionsetup{margin=6pt}
\caption{NCM training step}
\label{alg:ncm-training}
\scriptsize
\hypersetup{hypertexnames=false}
\hspace*{6pt}\begin{minipage}{\dimexpr\linewidth-12pt\relax}
\begin{algorithmic}[1]
\Require Text $x$, frozen tokenizer, $\theta$
\State Encode $x$ into target codes $r_{1:K}$
\State Uniformly replace codes to obtain $\tilde r_{1:K}$
\State $\tilde z^{(0)}\gets 0$
\For{$k=1,\ldots,K$}
    \State Resize and project $\tilde z^{(k-1)}$ to $\ell_k$ inputs
    \State $\tilde z^{(k)}\gets\tilde z^{(k-1)}+R_k(e_k(\tilde r_k))$
\EndFor
\State Predict all levels in one forward pass
\State Compute $\mathcal L_{\mathrm{NCM}}$ against $r_{1:K}$
\State Update $\theta$ using the loss gradient
\end{algorithmic}
\end{minipage}
\end{algorithm}
\end{minipage}\hfill
\begin{minipage}[t]{0.46\textwidth}
\begin{algorithm}[H]
\captionsetup{margin=6pt}
\caption{NCM generation}
\label{alg:ncm-generation}
\scriptsize
\hypersetup{hypertexnames=false}
\hspace*{6pt}\begin{minipage}{\dimexpr\linewidth-12pt\relax}
\begin{algorithmic}[1]
\Require Frozen tokenizer, generator $p_\theta$
\State $\hat z^{(0)}\gets 0$
\For{$k=1,\ldots,K$}
    \State Resize $\hat z^{(k-1)}$ to $\ell_k$ positions
    \State Project the resized vectors to the hidden width
    \State Predict code distributions in one forward pass
    \State Sample all $\ell_k$ codes $r_k^i$ in parallel
    \State $\hat z^{(k)}\gets\hat z^{(k-1)}+R_k(e_k(r_k))$
\EndFor
\State Decode $\hat z^{(K)}$ to token distributions
\State Select output tokens by argmax
\end{algorithmic}
\end{minipage}
\end{algorithm}
\end{minipage}}\par
\endgroup

\paragraph{Conditional generation.}
\name supports prompt-conditioned generation through a prefix-LM attention pattern. Prompt tokens are embedded using the frozen VQ-VAE encoder's token embedding table and projected into the hidden space. Prompt positions attend to each other; each target block attends to the prompt and itself, while prompt positions cannot attend to target blocks.

\newsavebox{\wrbox}
\newcommand{\XX}{\textcolor{cyan}{\textbf{XX}}}
\definecolor{oursrow}{HTML}{FFF1CC}
\clearpage
\section{Experiments}
\label{sec:results}

We evaluate \name\ along four axes:
\begin{itemize}
\item \textbf{Quality.} We first assess tokenizer quality through reconstruction accuracy and codebook utilization (\S\ref{subsec:tokenizer-analysis}). We then evaluate unconditional generation on TinyStories \citep{eldan2023tinystories}, One Billion Word (LM1B) \citep{chelba2013one}, and OpenWebText (OWT) \citep{Gokaslan2019OpenWeb}, measuring generative perplexity alongside sample diversity (\S\ref{subsec:uncond}). For conditional generation, we evaluate WMT14 German--English translation \citep{bojar2014wmt} using BLEU \citep{papineni2002bleu,sacrebleu} and XSum summarization \citep{narayan2018xsum} using ROUGE-1/2/L \citep{lin-2004-rouge} (\S\ref{subsec:seq2seq}).
\item \textbf{Efficiency.} We compare the number of function evaluations (NFE) and wall-clock time required to achieve a given generation quality, as well as the training-token budgets used to obtain these results (\S\ref{subsec:efficiency}).
\item \textbf{Reasoning.} We compare \name\ with autoregressive and diffusion baselines on Sudoku, where successful generation requires satisfying interdependent constraints across the entire output (\S\ref{subsec:reasoning}).
\item \textbf{Scaling.} We investigate how scaling the model size improves generation quality and whether \name's fast sampling can translate into further quality gains through additional test-time compute (\S\ref{subsec:scaling}).
\end{itemize}

Across these evaluations, \name\ achieves competitive generation quality with substantially lower sampling cost.
We train a separate tokenizer and generator for each dataset. Unless otherwise stated, our generators have approximately $100$M parameters, comparable to or smaller than the baselines. We follow established evaluation protocols from prior diffusion language modeling work where possible. Appendices~\ref{app:datasets}--\ref{app:extended_results} provide dataset, architecture, training, and baseline details, along with additional results.

\subsection{Tokenizer Quality}
\label{subsec:tokenizer-analysis}

\newcommand{\hd}[1]{\textbf{#1}}
\begin{table}[ht]
\centering
\small
\setlength{\tabcolsep}{4pt}
\renewcommand{\arraystretch}{1.12}
\begin{tabular*}{\linewidth}{@{\extracolsep{\fill}}l ccc cc ccc@{}}
\toprule
& \multicolumn{3}{c}{\hd{Unconditional}} & \multicolumn{2}{c}{\hd{Conditional}}
  & \multicolumn{3}{c}{\hd{Reasoning (Sudoku)}} \\
\cmidrule(lr){2-4}\cmidrule(lr){5-6}\cmidrule(l){7-9}
& OWT & LM1B & TinyStories & XSum & WMT14 & Easy & Medium & Hard \\
\midrule
Recon.\ acc.\ (\%)\,$\uparrow$ & 99.99 & 99.86 & 99.98 & 99.79 & 98.79 & 100.00 & 100.00 & 100.00 \\
Utilization (\%) & 83.10 & 89.31 & 90.12 & 86.20 & 61.89 & \phantom{0}80.73 & \phantom{0}83.85 & \phantom{0}80.73 \\
\bottomrule
\end{tabular*}
\vspace{3pt}
\caption{\textbf{Tokenizer reconstruction and codebook utilization.} Reconstruction accuracy measures exact token recovery on held-out text using the full hierarchy. Utilization is the mean percentage of codebook entries used across levels.}
\label{tab:recon}
\end{table}

{\brokenpenalty=10000
Reconstruction errors that are visually imperceptible in images can be consequential in text: e.g., one wrong digit can invalidate a Sudoku solution. Table~\ref{tab:recon} reports exact reconstruction accuracy. On Sudoku, where correctness requires exact recovery, the tokenizer achieves $100\%$ reconstruction accuracy. On other tasks, like summarization and translation, token-level reconstruction accuracy is close to $100\%$. Observed tokenizer reconstruction errors often involve substitutions of related words or entities, with their effect on meaning depending on the context. We also report tokenizer codebook utilization in Table~\ref{tab:recon}, with mean utilization ranging from $62\%$ to $90\%$ across datasets. Appendix~\ref{app:vqvae} provides further tokenizer analysis, including reconstruction errors and codebook utilization.
We provide additional experiments in Appendix~\ref{app:hierarchy} to further
understand the learned hierarchy.\par}

\subsection{Unconditional Generation}
\label{subsec:uncond}

\begin{figure}[!ht]
  \centering
  \includegraphics[width=\linewidth]{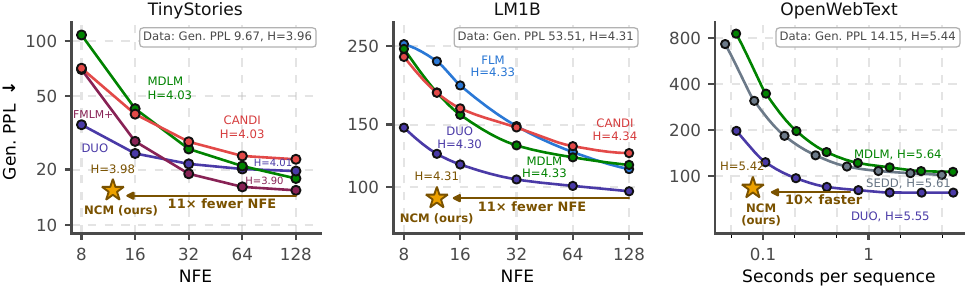}
  \caption{\textbf{Unconditional generation quality and sampling cost.} Generative perplexity (Gen.\ PPL; lower is better) against the number of function evaluations on TinyStories and LM1B, and against the sampling time per $1{,}024$-token sequence (batch size $16$) on OpenWebText. \name achieves competitive or better Gen.\ PPL compared to strong diffusion baselines with substantially fewer function evaluations and less time. $H$ denotes mean per-sample unigram entropy.}
  \label{fig:gen-ppl}
\end{figure}

Following standard practice in the diffusion language modeling literature \citep{sahoo2024mdlm,sahoo2025duo,zheng2025masked}, we evaluate \name's unconditional generation quality using generative perplexity (Gen.\ PPL), with GPT-2-large as the reference model on TinyStories \citep{eldan2023tinystories}, LM1B \citep{chelba2013one}, and OpenWebText (OWT) \citep{Gokaslan2019OpenWeb}. We report mean per-sample unigram entropy ($H$) alongside Gen.\ PPL, since lower Gen.\ PPL can be obtained by reducing sample diversity \citep{zheng2025masked}. For reference, we report both metrics on ground-truth text from each dataset. For a fair comparison, \name is compared against baselines with sample diversity close to that of the ground-truth data. Figure~\ref{fig:gen-ppl} compares \name with strong discrete and continuous diffusion language model baselines in terms of Gen.\ PPL at different numbers of function evaluations (NFE). Across the three datasets, \name achieves competitive or better Gen.\ PPL with substantially fewer function evaluations, while maintaining sample entropy close to that of the data. Table~\ref{tab:uncond-all} in Appendix~\ref{app:sweeps} lists all results with their entropies, and Appendix~\ref{app:ref-models} gives results under other reference scoring models.

\subsection{Conditional Generation}
\label{subsec:seq2seq}

We evaluate \name on translation (WMT14 German--English, De--En) \citep{bojar2014wmt} and extreme summarization (XSum) \citep{narayan2018xsum}, using BLEU \citep{papineni2002bleu,sacrebleu} and ROUGE-1/2/L \citep{lin-2004-rouge} scores, respectively. We compare \name with autoregressive baselines \citep{hu2026elf} and diffusion and flow models \citep{sahoo2024mdlm,sahoo2025duo,arriola2025e2d2,yuan2022seqdiffuseq,hu2026elf,liang2026aurora} at comparable model sizes (Table~\ref{tab:seq2seq}). Without guidance, \name achieves better ROUGE-1/2/L scores on XSum than the strongest diffusion baseline, AURORA-LM \citep{liang2026aurora}, while reducing the number of function evaluations (NFE) by $16\times$. Among the non-AR baselines on WMT14, \name trails only the strongest diffusion baseline, E2D2 \citep{arriola2025e2d2}, while using $5.8\times$ fewer NFE. \name's lower sampling cost allows us to match the baselines' NFE budget using multiple candidates and minimum Bayes risk (MBR) decoding. \name with MBR achieves better scores on all three XSum ROUGE metrics than the strongest diffusion baseline with guidance, AURORA-LM \citep{liang2026aurora}, while using fewer NFE (Table~\ref{tab:seq2seq}). We count each classifier-free guidance step as two function evaluations, since it requires both a conditional and an unconditional generator call. We visualize generated samples in Figure~\ref{fig:samples} and Appendix~\ref{sec:qualitative-extra}.

\begin{table}[thb]
\centering
\footnotesize
\setlength{\tabcolsep}{3pt}
\renewcommand{\arraystretch}{1.05}
\begin{tabular}{@{}l l ccc ccccc@{}}
\toprule
& & \multicolumn{3}{c}{\textbf{WMT14 De--En}}
  & \multicolumn{5}{c}{\textbf{XSum}} \\
\cmidrule(lr){3-5}\cmidrule(l){6-10}
\textbf{Type} & \textbf{Method} & \textbf{Size}
  & BLEU\,$\uparrow$ & NFE\,$\downarrow$
  & \textbf{Size} & {\scriptsize ROUGE-1\,$\uparrow$} & {\scriptsize ROUGE-2\,$\uparrow$} & {\scriptsize ROUGE-L\,$\uparrow$} & NFE\,$\downarrow$ \\
\midrule
  \textcolor{black!55}{--} & \textcolor{black!55}{AR\rlap{$^\dagger$}} & \textcolor{black!55}{99M}
    & \textcolor{black!55}{25.2} & \textcolor{black!55}{64}
    & \textcolor{black!55}{99M} & \textcolor{black!55}{30.5} & \textcolor{black!55}{10.2}
    & \textcolor{black!55}{24.4} & \textcolor{black!55}{64} \\
\midrule
\multicolumn{10}{@{}l}{\itshape\color{black!55} Without guidance} \\[1pt]
\midrule
  \multirow{4}{*}{Diff.} & MDLM\rlap{$^\dagger$} & 99M
    & 18.4 & \underline{64} & 99M & 33.4 & 11.6 & 25.8 & \underline{64} \\
         & DUO\rlap{$^\dagger$} & 205M
    & 21.3 & \underline{64} & 205M & 31.4 & 10.1 & 25.0 & \underline{64} \\
         & E2D2\rlap{$^\dagger$} & 99M
    & \textbf{24.8} & \underline{64} & 99M & 28.4 & 8.3 & 22.0 & \underline{64} \\
         & SeqDiffuSeq\rlap{$^\dagger$} & --
    & 21.3 & \underline{64} & -- & 19.3 & 1.7 & 14.1 & \underline{64} \\
\midrule
  \multirow{2}{*}{Flow} & ELF & 140M
    & 17.2 & \underline{64} & 140M & 34.2 & 11.1 & 26.4 & \underline{64} \\
         & AURORA-LM & --
    & -- & -- & 130M & \underline{34.7} & \underline{12.2} & \underline{26.6} & 128 \\
\midrule
  \rowcolor{oursrow} Ours & \name & 120M
    & \underline{22.6} & \textbf{11}
    & 88M & \textbf{35.0} & \textbf{12.8} & \textbf{28.0} & \textbf{8} \\
\midrule[\heavyrulewidth]
\multicolumn{10}{@{}l}{\itshape\color{black!55} With guidance or MBR decoding} \\[1pt]
\midrule
  \multirow{2}{*}{Flow} & ELF & 140M
    & \textbf{26.4} & \underline{128} & 140M & 36.0 & 12.2 & 27.8 & \underline{128} \\
         & AURORA-LM & --
    & -- & -- & 130M & \underline{36.3} & \underline{13.4} & \underline{28.6} & 184 \\
\midrule
  \rowcolor{oursrow} & \name + CFG & 120M
    & 23.9 & \textbf{22} & 88M & 35.3 & 13.0 & 28.2 & \textbf{16} \\
  \rowcolor{oursrow} \multirow{-2}{*}{Ours} & \name + MBR & 120M
    & \underline{25.2} & 132 & 88M & \textbf{37.3} & \textbf{14.0} & \textbf{29.6} & \underline{128} \\
\bottomrule
\end{tabular}
\vspace{3pt}
\caption{\textbf{Conditional generation quality and sampling cost.}
BLEU on WMT14 De--En and ROUGE-1/2/L on XSum (higher is better). Size denotes task-specific generator parameters; NFE counts generator calls per output. Bold and underlined values mark the best and second-best results within each non-AR decoding setting, including ties. $^\dagger$: results from \citet{hu2026elf}.}
\label{tab:seq2seq}
\end{table}

\Needspace*{15\baselineskip}
\subsection{Reasoning}
\label{subsec:reasoning}

\sbox{\wrbox}{\footnotesize\setlength{\tabcolsep}{3pt}\renewcommand{\arraystretch}{1.05}\begin{tabular}{@{}l l c c c c c@{}}
  \toprule
  & & & & \multicolumn{3}{c@{}}{\textbf{Accuracy (\%)}\,$\uparrow$} \\
  \cmidrule(l){5-7}
  \textbf{Type} & \textbf{Method} & \textbf{Size} & \textbf{NFE}\,$\downarrow$ & Easy & Med. & Hard \\
  \midrule
  \multirow{2}{*}{AR}    & Sample & 25M & 128 & 13.9 & 5.1  & 0.6  \\
                         & Greedy & 25M & 128 & 14.6 & 5.1  & 1.0  \\
  \midrule
  \multirow{2}{*}{Disc.} & MDLM   & 29M & 128 & 92.0 & 77.1 & 30.2 \\
                         & DUO    & 29M & 128 & 96.3 & 84.7 & 58.4 \\
  \midrule
  \multirow{4}{*}{Cont.} & CANDI  & --  & 128 & 79.3 & 45.9 & 16.7 \\
                         & FLM    & 29M & 128 & 94.2 & 82.7 & 44.5 \\
                         & S-FLM  & 29M & 128 & 94.8 & 85.2 & 45.0 \\
                         & FMLM+  & 29M & 16  & 97.7 & 92.8 & 81.3 \\
  \midrule
  \rowcolor{oursrow} Ours & \name & 25M & \textbf{12}  & \textbf{99.5} & \textbf{96.5} & \textbf{86.1} \\
  \bottomrule
  \end{tabular}}\begin{wraptable}{r}{\wd\wrbox}
  \vspace{-\intextsep}
  \centering
  \usebox{\wrbox}
  \vspace{2pt}
  \captionsetup{justification=justified,singlelinecheck=false}
  \caption{\textbf{Sudoku.} Completely solved boards (\%) by difficulty. Baseline results: \citet{agarwal2026posterior}. \name outperforms all baselines with fewer NFE.}
  \label{tab:reasoning}
\end{wraptable}

We evaluate \name on Sudoku, a reasoning task in which every cell of the solution must satisfy the row, column, and box constraints at once. Following the setup of S-FLM \citep{deschenaux2026language}, we report the percentage of completely solved boards at three difficulty levels defined by the number of provided clues ($40$, $35$, and $30$ for easy, medium, and hard, respectively). \name outperforms both the AR baselines and the strongest diffusion baseline, FMLM+ \citep{agarwal2026posterior}, across all three levels of difficulty while using fewer NFE (Table~\ref{tab:reasoning}). \name's coarse-to-fine generation is well suited to tasks that require global planning: \name refines a representation of the entire board across levels, allowing later levels to correct earlier partial solutions. In contrast, the AR baselines struggle with left-to-right generation, which commits to individual tokens without revising earlier decisions. Appendix~\ref{app:sudoku-levels} shows a sample Sudoku solving trajectory across different time steps.

\subsection{Scaling}
\label{subsec:scaling}

We study how \name benefits from additional compute through larger models and increased inference budgets. We evaluate model scaling by increasing generator size and test-time scaling by selecting among multiple candidates using minimum Bayes risk (MBR) decoding.

\paragraph{Model Scaling.}

\begin{figure}[!t]
\centering
\begin{subfigure}[t]{0.315\linewidth}
  \centering
  \includegraphics[width=\linewidth]{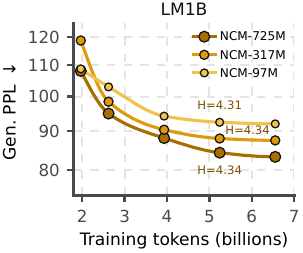}
  \caption{Model Scaling}
  \label{fig:scaling}
\end{subfigure}\hfill
\begin{subfigure}[t]{0.655\linewidth}
  \centering
  \includegraphics[width=0.48\linewidth]{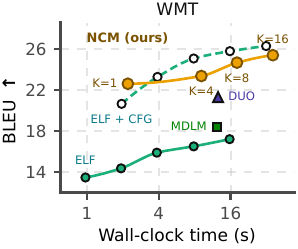}\hfill
  \includegraphics[width=0.48\linewidth]{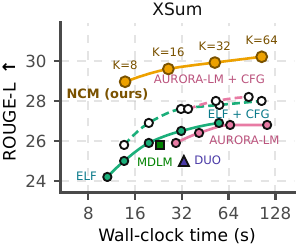}
  \caption{Test-time Scaling}
  \label{fig:tts-wmt}
  \label{fig:tts}
\end{subfigure}
\caption{\textbf{Model and test-time scaling.}
\textbf{(a)}~Increasing generator size improves Gen.\ PPL while maintaining high sample entropy ($H$) on LM1B at a fixed generator training-token budget; the shared tokenizer is excluded from this panel.
\textbf{(b)}~Generating more candidates and selecting with MBR decoding improves BLEU on WMT14 and ROUGE-L on XSum. $K$ denotes the number of candidates; wall-clock time measures sampling cost. On XSum, \name outperforms diffusion baselines at matched inference time.}
\label{fig:efficiency-group}
\end{figure}

Scaling \name's model size improves both generation quality and sample diversity at a fixed training-token budget (Figure~\ref{fig:scaling}). We train generators with 97M, 317M, and 725M parameters on LM1B, all sharing the same frozen tokenizer. These results suggest that \name scales effectively, with potential for further gains from larger generators.

\paragraph{Test-Time Scaling.}
\label{subsubsec:tts}

In Figure~\ref{fig:efficiency-latency} and Table~\ref{tab:seq2seq}, we show that \name matches or outperforms diffusion baselines while using significantly fewer function evaluations (NFEs) and less wall-clock time. A natural question is whether we can convert these efficiency gains into further improvements in quality. We generate multiple candidates and use minimum Bayes risk (MBR) decoding \citep{bertsch2023mbr} to select the candidate with the highest average similarity to the others. On XSum, Figure~\ref{fig:tts} shows that \name improves over diffusion baselines at matched inference time.

\subsection{Efficiency}
\label{subsec:efficiency}

\paragraph{Inference Efficiency.}
\label{subsubsec:inference-eff}
In Figure~\ref{fig:gen-ppl}, we show that \name achieves lower or comparable generative perplexity than the strongest diffusion baselines using far fewer function evaluations (NFE). However, NFE alone does not capture the full inference efficiency picture, since the cost per step differs across different diffusion language models; many DLMs process the full sequence at every step and use samplers that can introduce additional overhead. \name, on the other hand, processes very few tokens at the coarse levels of its hierarchy, saving compute at these early steps. We benchmark end-to-end wall-clock time for generating $1{,}024$ sequences at batch size $16$ on a single NVIDIA H100 SXM 80GB GPU (Figure~\ref{fig:efficiency-latency}). On LM1B, \name uses $11\times$ fewer NFE than the strongest diffusion language model baseline, DUO, and runs $21\times$ faster at better generative perplexity. On XSum, it uses $16\times$ fewer NFE than the strongest diffusion baseline, AURORA-LM, and runs $31\times$ faster at comparable ROUGE. The speedup gains come from both fewer NFE and a lower average cost per step compared to the DLM baselines.

\par
\begingroup\setlength{\parfillskip}{0pt}\noindent
\begin{minipage}[c]{0.52\linewidth}
\paragraph{Lower Training Budgets.}
\label{subsubsec:train-eff}
\name\ achieves strong unconditional generation results with fewer training tokens than the diffusion baselines. On LM1B, the tokenizer and generator together process $13.36$B full-length token positions, approximately $4.9\times$ fewer than the baselines' $66$B (Figure~\ref{fig:lm1b_tokens}). We count repeated passes over the data in this budget; the calculation is given in Appendix~\ref{app:training-token-accounting}. These budgets measure total token exposure. One possible explanation is that \name's fixed coarse-to-fine order simplifies learning relative to the many generation orders learned by diffusion baselines.
\end{minipage}\hspace{0.03\linewidth}\begin{minipage}[c]{0.45\linewidth}
  \centering
  \begin{tikzpicture}
  \begin{axis}[
      width=0.97\linewidth, height=4.8cm,
      ybar, bar width=10pt, bar shift=0pt,
      ymin=0, ymax=80,
      xmin=0.4, xmax=6.6,
      xtick={1,...,6},
      xticklabels={\name, RDLM, CANDI, MDLM, DUO, FLM},
      xticklabel style={font=\tiny},
      yticklabel style={font=\scriptsize},
      ytick={0,20,40,60},
      ylabel={Training tokens (B)},
      ylabel style={font=\small},
      ymajorgrids, grid style={dashed, gray!30},
      axis x line*=bottom, axis y line*=left,
      axis line style={gray!70!black, line width=0.8pt},
      point meta=explicit symbolic,
      nodes near coords={\pgfplotspointmeta},
      every node near coord/.append style={font=\tiny, inner sep=1.5pt},
      x tick style={draw=none},
  ]
    \addplot[fill=cNCM,  draw=cNCMtxt, line width=0.4pt,
             every node near coord/.append style={font=\tiny\bfseries, text=cNCMtxt}]
             coordinates {(1,13.36) [93.1]};
    \addplot[fill=cRDLM, draw=black, line width=0.4pt,
             every node near coord/.append style={text=cRDLMtxt}] coordinates {(2,66) [295.4]};
    \addplot[fill=cCANDI, draw=black, line width=0.4pt,
             every node near coord/.append style={text=cCANDItxt}] coordinates {(3,66) [124.6]};
    \addplot[fill=cMDLM, draw=black, line width=0.4pt,
             every node near coord/.append style={text=cMDLMtxt}] coordinates {(4,66) [115.5]};
    \addplot[fill=cDUO, draw=black, line width=0.4pt,
             every node near coord/.append style={text=cDUOtxt}] coordinates {(5,66) [97.3]};
    \addplot[fill=cFLM, draw=black, line width=0.4pt,
             every node near coord/.append style={text=cFLMtxt}] coordinates {(6,66) [112.3]};
    \node[font=\scriptsize\bfseries, text=cNCMtxt, align=center]
         (callout) at (axis cs:1,42) {20\%\\ tokens};
    \draw[->, cNCMtxt, line width=0.7pt] (callout.south) -- (axis cs:1,21);
  \end{axis}
  \end{tikzpicture}
  \captionsetup{justification=raggedright,singlelinecheck=false}
  
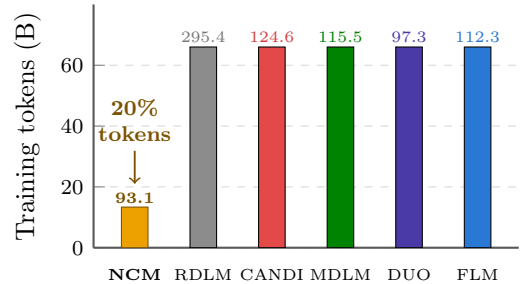
\captionof{figure}{\textbf{Training tokens on LM1B.} \name\ uses $4.9\times$ fewer tokens and
  reaches a lower Gen.\ PPL (above each bar).}
  \label{fig:lm1b_tokens}
\end{minipage}
\par\endgroup

\subsection{Qualitative Samples}
\label{sec:qualitative}
Figure~\ref{fig:samples} shows \name samples across unconditional and conditional settings. The conditional outputs remain anchored to their input contexts. Additional samples are provided in Appendix~\ref{sec:qualitative-extra}.

\newlength{\smplbl}
\tcbset{samplestyle/.style={
  enhanced,
  arc=0pt,
  boxrule=0.8pt,
  left=2.5mm, right=2.5mm, top=1.8mm, bottom=1.8mm,
  before skip=5pt, after skip=7pt,
}}

\newcommand{\splabel}[2]{\textbf{#2}}
\newcommand{\spheading}[3]{{\footnotesize\textbf{#2}}\hfill{\scriptsize #3}\par
  \vspace{1pt}{\color{black}\hrule height 0.8pt}\vspace{4pt}}
\newcommand{\sprows}[3]{\settowidth{\smplbl}{\textbf{Generated:}}\begin{tabular}{@{}l@{\hspace{6pt}}p{\dimexpr\linewidth-\smplbl-6pt\relax}@{}}
  \splabel{spblue}{Reference}\textbf{:} & #2 \\[2pt]
  \splabel{#1}{Generated}\textbf{:} & #3 \\
  \end{tabular}}
\newcommand{\sprule}[1]{\vspace{3.5pt}{\color{black!35}\hrule height 0.4pt}\vspace{3.5pt}}

\definecolor{uncbg}{RGB}{240,250,240}
\definecolor{uncframe}{RGB}{34,139,34}
\definecolor{condbg}{RGB}{254,249,235}
\definecolor{condframe}{RGB}{176,131,20}
\definecolor{sumbg}{RGB}{253,244,243}
\definecolor{sumframe}{RGB}{190,60,55}
\definecolor{transbg}{RGB}{238,246,253}
\definecolor{promptblue}{RGB}{0,51,204}
\definecolor{spblue}{RGB}{13,110,180}
\definecolor{transframe}{RGB}{13,110,180}

\begin{figure}[H]
\tcbset{samplestyle/.append style={top=1mm,bottom=1mm,before skip=0pt,after skip=3pt}}
\renewcommand{\spheading}[3]{{\footnotesize\textbf{#2}}\hfill{\scriptsize #3}\par
  \vspace{1.5pt}{\color{black}\hrule height 0.8pt}\vspace{4pt}}
\renewcommand{\sprule}[1]{\vspace{3.5pt}{\color{black!35}\hrule height 0.4pt}\vspace{3.5pt}}

\begin{tcolorbox}[samplestyle, colback=white, colframe=black]
\scriptsize\linespread{1}\selectfont
\spheading{spblue}{Summarization}{\textbf{ROUGE-1:} 48.5\;\textcolor{black!25}{\textbar}\; \textbf{ROUGE-2:}
  12.9\;\textcolor{black!25}{\textbar}\; \textbf{ROUGE-L:} 36.4}
\noindent\splabel{spblue}{Context}\textbf{:}\;
The 33-year-old has featured only twice for the Foxes this
term, having signed a new one-year deal with the Premier League newcomers in the
summer. Former Blackpool forward Taylor-Fletcher scored three goals in 23 games
for his parent club last season. He joined the 13th-placed Owls prior to
Saturday's Championship game against Norwich City.
\sprule{spblue}
\sprows{spblue}{Leicester City striker Gary Taylor-Fletcher has joined
  Sheffield Wednesday on an initial month-long loan.}{Sheffield Wednesday have
  signed Leicester striker James Taylor-Fletcher on loan until the end of the
  season.}
\end{tcolorbox}

\begin{tcolorbox}[samplestyle, colback=white, colframe=black]
\scriptsize\linespread{1}\selectfont
\spheading{spblue}{Translation}{\textbf{BLEU:} 51.9}
\settowidth{\smplbl}{\textbf{Generated:}}\begin{tabular}{@{}l@{\hspace{6pt}}p{\dimexpr\linewidth-\smplbl-6pt\relax}@{}}
\splabel{spblue}{German}\textbf{:} &
  Leider gibt es kein Wundermittel zur Vorbeugung von Krebs. \\
\end{tabular}
\sprule{spblue}
\sprows{spblue}{Unfortunately, there is no miracle recipe for preventing cancer.}{Unfortunately, there is no miracle cure to prevent cancer.}
\end{tcolorbox}

\begin{tcolorbox}[samplestyle, colback=white, colframe=black]
\scriptsize\linespread{1}\selectfont
\spheading{spblue}{Conditional Generation}{\textcolor{promptblue}{Prompt}\;\textcolor{black!25}{\textbar}\;
  Generated}
\noindent\splabel{spblue}{Sample 1}\textbf{:}\;
{\color{promptblue}An Egyptian court has ordered a probe into fresh clashes
between police and protesters in Cairo. ``There are strikes at the Suez
Canal,''.} witnesses said. ``It's going to be happening in
the streets. It's the new regime.'' Those police, responding to the protests,
are now getting the money out of the government, he said\ldots

\par\vspace{5pt}
\noindent\splabel{spblue}{Sample 2}\textbf{:}\;
But Church explosion and billions of thought, frustration, morals, and religion had itself bound inextricably in the fabric of beauty rather than reason. The decay of {\color{promptblue}society fell into the wealth of divine advancement at hand. The power of social engineering perused human godhood into a sense of direction.}
\end{tcolorbox}

\begin{tcolorbox}[samplestyle, colback=white, colframe=black]
\scriptsize\linespread{1}\selectfont
\spheading{spblue}{Unconditional Generation}{\textbf{Gen.\ PPL:} 55.6\;\textcolor{black!25}{\textbar}\; \textbf{Entropy:}
  4.39}
\noindent \textbf{Generated:} {[}CLS] at the far end of the series around 18 games. {[}CLS]
\textquotedbl{}it could be good if he doesn\textquotesingle{}t simply name the
party that has won his past two on the other side and the corners,
\textquotedbl{} he said. {[}CLS] \textquotedbl{}if you are in power then you can
put your players down and work together and try find out the changes with which
we need. {[}CLS] i work in the various schools or into
children\textquotesingle{}s school. {[}CLS] saunders, who was also hit after
britain\textquotesingle{}s most successful player in 30 years, had been for six
appearances since starting 11 with an 11--14 major league record, a 28-year high
{[}CLS]
\end{tcolorbox}
\captionsetup{font=small,aboveskip=6pt}
\caption{\textbf{Qualitative samples from \name.} Unconditional generation samples are on LM1B. Conditional generation samples use prompts from a held-out OWT validation set. Summarization and translation samples are on XSum and WMT14 De--En, respectively. Additional samples are provided in Appendix~\ref{sec:qualitative-extra}.}
\label{fig:samples}
\end{figure}

\clearpage

\section{Conclusion}
\label{sec:conc}

We introduce \textbf{N}ext-\textbf{C}oncept \textbf{M}odeling (\name), a
two-stage framework that first learns a discrete coarse-to-fine representation
of text and then trains a transformer to generate its levels in order.
Predicting all codes within each level in parallel provides a short sequential
generation path. On unconditional generation across TinyStories, LM1B, and
OpenWebText, \name achieves competitive or better Gen.\ PPL than strong diffusion
baselines with approximately $4$--$11\times$ fewer function evaluations (NFE),
while maintaining sample entropy close to the data. Generation is
$10$--$21\times$ faster on LM1B and OpenWebText. On conditional generation,
\name uses approximately $6$--$16\times$ fewer NFE than unguided diffusion
baselines on WMT14 translation and XSum summarization, achieving competitive
translation quality and the best summarization scores. On XSum, it runs
$31\times$ faster than the strongest diffusion baseline. It also uses
approximately $4$--$5\times$ fewer training tokens than strong diffusion
baselines on LM1B and OpenWebText, including both tokenizer and generator
training. Finally, preliminary scaling results on LM1B show that increasing
the generator size improves generation quality at a fixed training-token budget.

\section*{Acknowledgments}
\label{sec:ack}
We thank Yasa Baig, Dan Biderman, Bradley Brown, Francois Chaubard, Mayee Chen, Vivien Cheng, Catherine Deng, Owen Dugan, Sabri Eyuboglu, Simon Guo, Neel Guha, Junmiao Hu, Hangoo Kang, Jacky Kwok, Jerry Liu, Avanika Narayan, Divya Nori, Jon Saad-Falcon, Ben Spector, Stuart Sul, Tarun Suresh, Shayan Talaei, Roberto Torres, Ateh Wangia, and Michael Zhang for their helpful feedback and discussions.

\vspace{1\baselineskip}

\noindent We gratefully acknowledge the support of NIH under Nos. U54EB020405 (Mobilize) and R01GM124443 (SimTK); NSF under Nos. CCF2247015 (Hardware-Aware), CCF1763315 (Beyond Sparsity), CCF1563078 (Volume to Velocity), 1937301 (RTML), 1900638 (Open Federated Virtual Assistants), and 2028602 (PPoSS); DARPA under No. HR00112520038 (Fallingwater); US DEVCOM ARL under Nos. W911NF-23-2-0184 (Long-context) and W911NF-21-2-0251 (Interactive Human-AI Teaming); NSTXL under No. N00164-23-9-G057-01 (Energy-Efficient AI Hardware); ONR under Nos. N000142312633 (Deep Signal Processing), N000142212426 (Misinformation Analysis), N000142012480 (Non-Euclidean Geometry), and N000142012275 (Data Programming); Stanford HAI under Nos. 247183, 215955, and 345077 (Evo); Google DeepMind; Google Research; Google Cloud; NXP, Xilinx, LETI-CEA, Intel, IBM, Microsoft, NEC, Toshiba, TSMC, ARM, Hitachi, BASF, Accenture, Ericsson, Qualcomm, Analog Devices, Salesforce, Total, the Laude Institute, Prime Intellect, Anthropic, the HAI-GCP Cloud Credits for Research program, the Stanford Data Science Initiative (SDSI), the Stanford Marlowe Computing Platform, Lambda Labs, Modal, Zoltan's Flops, and members of the Stanford DAWN project: Meta, Google, and VMWare; and members of the Stanford SEAMS project: IBM and Felicis. HK is supported by the Knight Hennessy Scholars Program. IK is supported in part by the NSF GRFP. The U.S. Government is authorized to reproduce and distribute reprints for Governmental purposes notwithstanding any copyright notation thereon. Any opinions, findings, and conclusions or recommendations expressed in this material are those of the authors and do not necessarily reflect the views, policies, or endorsements, either expressed or implied, of NIH, ONR, DARPA, or the U.S Government.

\clearpage
\bibliographystyle{unsrtnat}
\bibliography{main}

\clearpage
\appendix

\startcontents[appendices]
\section*{Appendix Contents}
\printcontents[appendices]{}{0}{\setcounter{tocdepth}{2}}
\clearpage

\section{Probabilistic Formulation and ELBO}
\label{app:theory}

We formulate \name\ as a hierarchical latent-variable model and derive an evidence
lower bound (ELBO) on its text log-likelihood. We consider unconditional generation
with deterministic nearest-neighbor code assignments.

\subsection{Setup}
\label{app:theory:setup}

Following Section~\ref{sec:method}, let $x\in\mathcal T^L$ be a sequence of $L$ tokens
from the text vocabulary $\mathcal T$. The tokenizer assigns $x$ a hierarchy
$r^*_{1:K}(x)=(r^*_1,\ldots,r^*_K)$, where
$r_k^*\in\{1,\ldots,|\mathcal V_k|\}^{\ell_k}$ and
$\ell_1<\cdots<\ell_K=L$. We fix the tokenizer's parameters and use a fixed
rule to break nearest-neighbor ties, so this assignment is deterministic.
For any code hierarchy $r_{1:K}$, define its cumulative reconstruction by
\begin{equation}
  \hat z^{(k)}(r_{1:k})=\sum_{j=1}^{k}R_j\bigl(e_j(r_j)\bigr),
  \qquad \hat z^{(0)}=0,
  \label{eq:thy:recon}
\end{equation}
where $e_j$ is the codebook lookup and $R_j$ is the reconstruction operator
of Section~\ref{subsec:tokenizer}.
The decoder defines a likelihood over text,
\begin{equation}
  p_\psi(x\mid r_{1:K})
  =\prod_{n=1}^{L}p_\psi\bigl(x^n\mid\hat z^{(K)}(r_{1:K})\bigr),
  \label{eq:thy:decoder}
\end{equation}
and the generator defines a prior over hierarchies,
\begin{equation}
  p_\theta(r_{1:K})
  =\prod_{k=1}^{K}p_\theta\bigl(r_k\mid\hat z^{(k-1)}\bigr)
  =\prod_{k=1}^{K}\prod_{i=1}^{\ell_k}
    p_\theta\bigl(r_k^i\mid\hat z^{(k-1)}\bigr),
  \label{eq:thy:prior}
\end{equation}
where $\hat z^{(k-1)}=\hat z^{(k-1)}(r_{1:k-1})$. Each factor is normalized
for every preceding hierarchy, so their product defines a normalized prior.
The joint distribution and marginal text likelihood are
\begin{equation}
  p_{\theta,\psi}(x,r_{1:K})=p_\psi(x\mid r_{1:K})p_\theta(r_{1:K}),
  \qquad
  p_{\theta,\psi}(x)=\sum_{r_{1:K}}p_\psi(x\mid r_{1:K})p_\theta(r_{1:K}).
  \label{eq:thy:joint}
\end{equation}

\subsection{Evidence Lower Bound}
\label{app:theory:elbo}

Use the tokenizer's deterministic assignment as a point-mass variational distribution,
\begin{equation}
  q(r_{1:K}\mid x)=\mathbf 1\{r_{1:K}=r^*_{1:K}(x)\}.
\end{equation}
This distribution has zero entropy. Its ELBO therefore reduces to the log joint
probability of the tokenizer's assigned hierarchy:
\begin{equation}
  \begin{aligned}
  \mathcal L_{\mathrm{ELBO}}(x)
  &\equiv \mathbb E_{q(r_{1:K}\mid x)}
    \left[\log p_{\theta,\psi}(x,r_{1:K})-\log q(r_{1:K}\mid x)\right]\\
  &=\log p_\psi(x\mid r^*_{1:K})+\log p_\theta(r^*_{1:K})\\
  &\le\log\sum_{r_{1:K}}p_\psi(x\mid r_{1:K})p_\theta(r_{1:K})
   =\log p_{\theta,\psi}(x).
  \end{aligned}
  \label{eq:thy:bound}
\end{equation}
The inequality follows because the marginal likelihood sums nonnegative terms and
includes the term for $r^*_{1:K}$. Writing
$\hat z_*^{(k)}=\hat z^{(k)}(r^*_{1:k})$ and substituting
\eqref{eq:thy:decoder} and~\eqref{eq:thy:prior} gives
\begin{equation}
  \boxed{
  \mathcal L_{\mathrm{ELBO}}(x)=
  \underbrace{\sum_{n=1}^{L}\log p_\psi\bigl(x^n\mid\hat z_*^{(K)}\bigr)}_{\text{reconstruction}}
  +
  \underbrace{\sum_{k=1}^{K}\sum_{i=1}^{\ell_k}
    \log p_\theta\bigl(r_k^{*i}\mid\hat z_*^{(k-1)}\bigr)}_{\text{generation}}
  }
  \label{eq:thy:elbo}
\end{equation}
where $r_k^{*i}$ is the $i$th code at level $k$ in the assigned hierarchy.
The gap to the log-likelihood is exactly
\begin{equation}
  \begin{aligned}
  \log p_{\theta,\psi}(x)-\mathcal L_{\mathrm{ELBO}}(x)
  &=D_{\mathrm{KL}}\!\left(q(r_{1:K}\mid x)\,\|\,
      p_{\theta,\psi}(r_{1:K}\mid x)\right)\\
  &=-\log p_{\theta,\psi}(r^*_{1:K}\mid x)\ge0.
  \end{aligned}
  \label{eq:thy:gap}
\end{equation}
Thus, the bound is tight if and only if the posterior assigns probability one to
the tokenizer's hierarchy. With finite softmax logits, the prior and decoder assign
positive probability to every hierarchy and text sequence, so the bound and gap
are finite.

For a frozen tokenizer and decoder, only the generation term depends on $\theta$.
With clean conditioning and $\alpha=0$, the generator loss in
Section~\ref{subsec:training} is the negative generation term divided by the total
number of codes $N=\sum_{k=1}^{K}\ell_k$; minimizing it therefore maximizes this
ELBO over $\theta$. Conditioning corruption and nonuniform level weighting modify
that objective. The bound uses the complete hierarchy; nested dropout and the
quantization losses used to train the tokenizer are not included in it.

\section{Datasets}
\label{app:datasets}
\label{app:data}

We train a separate tokenizer and generator for each of six datasets, grouped by task
as in Section~\ref{sec:results}. Table~\ref{tab:datasets} summarizes them.

\begin{table}[ht]
\centering
\footnotesize
\setlength{\tabcolsep}{3.5pt}
\renewcommand{\arraystretch}{1.1}
\begin{tabularx}{\linewidth}{@{}>{\raggedright\arraybackslash}X l l l l c@{}}
\toprule
\textbf{Dataset} & \textbf{Abbreviation} & \textbf{Task} & \textbf{Text tokenizer} & \textbf{Context $c$}
  & \textbf{Target $L$} \\
\midrule
OpenWebText~\citep{Gokaslan2019OpenWeb} & OWT & Unconditional & GPT-2 & Null prefix & 1{,}024 \\
One Billion Word~\citep{chelba2013one} & LM1B & Unconditional & \texttt{bert-base-uncased} & Null prefix & 128 \\
TinyStories~\citep{eldan2023tinystories} & -- & Unconditional & Custom BPE & Null prefix & 128 \\
\midrule
Extreme Summarization~\citep{narayan2018xsum} & XSum & Summarization & GPT-2 & Article (512) & 48 \\
WMT 2014 German--English~\citep{bojar2014wmt} & WMT14 & Translation & \texttt{Qwen3-0.6B} & German (96) & 64 \\
\midrule
Sudoku~\citep{deschenaux2026language} & -- & Reasoning & FMLM+~\citep{agarwal2026posterior} & Puzzle & 89 \\
\bottomrule
\end{tabularx}
\caption{\textbf{Datasets.} Target length $L$ is the number of target tokens, which equals
the length $\ell_K$ of the finest level; context lengths are in tokens. ``--'' indicates
that we use the full dataset name rather than a separate abbreviation.}
\label{tab:datasets}
\end{table}

\section{Architecture Details}
\label{app:architecture}
\label{app:extended-architecture}
\label{app:training}
\label{app:impl}

A tokenizer and a generator are trained for each dataset (Appendix~\ref{app:datasets}):
the tokenizer first, and it is then frozen as the generator trains. For XSum and WMT14,
the generator is initially trained with unpaired text, then with source--target pairs.
The target length schedule and level schedule remain unchanged.
Table~\ref{tab:tokenizer-config} describes the tokenizers' architectures and training.
Table~\ref{tab:generator-config} provides details about the generators' architectures,
training, and sampling. The datasets are listed in the same order as in
Table~\ref{tab:recon}. All experiments employed seed 42, BF16 mixed-precision and
\texttt{torch.compile}.

\newcommand{\cfgrow}[1]{\multicolumn{7}{@{}l}{\itshape\color{black!55} #1} \\[1pt]}

\Needspace*{34\baselineskip}
\subsection{Multiscale VQVAE Tokenizer}
\label{app:tokenizer-training}

\begin{table}[H]
\centering
\footnotesize
\setlength{\tabcolsep}{3.6pt}
\renewcommand{\arraystretch}{1.08}
\begin{tabular}{@{}l ccc cc c@{}}
\toprule
\textbf{Setting} & \textbf{OWT} & \textbf{LM1B} & \textbf{TinyStories} & \textbf{XSum} & \textbf{WMT14} & \textbf{Sudoku} \\
\midrule
\cfgrow{Architecture}
Text vocab.\ size                 & 50,304 & 30,528 & 8,192 & 50,304 & 151,680 & 12 \\
Target length $L$                 & 1,024 & 128 & 128 & 48 & 64 & 89 \\
Levels $K$                        & 32 & 12 & 12 & 8 & 11 & 12 \\
Codebook vocab.\ $V_k$            & 4,096 & 8,192 & 4,096 & 4,096 & 4,096 & 16 \\
Latent width $d$                  & 384 & 384 & 384 & 384 & 384 & 256 \\
Encoder blocks                    & 0 & 0 & 0 & 0 & 0 & 0 \\
Decoder blocks                    & 1 & 4 & 1 & 1 & 1 & 1 \\
Decoder attn.\ heads              & 6 & 6 & 6 & 6 & 6 & 4 \\
Dropout                           & 0.3 & 0.1 & 0.3 & 0.1 & 0.1 & 0.3 \\
Parameters                        & 111.2M & 111.40M & 36.43M & 60.63M & 144.00M & 14.90M \\
\midrule
\cfgrow{Training}
Training updates                  & 78,432 & 25,957 & 30,000 & 120,001 & 60,000 & 10,000 \\
Global batch size                 & 768 & 2,048 & 4,096 & 4,096 & 4,096 & 1,024 \\
Peak learning rate                & $3\times10^{-4}$ & $3\times10^{-4}$ & $5\times10^{-4}$ & $5\times10^{-4}$ & $5\times10^{-4}$ & $5\times10^{-4}$ \\
Learning-rate floor               & $3\times10^{-5}$ & $3\times10^{-5}$ & $5\times10^{-5}$ & $5\times10^{-5}$ & $5\times10^{-5}$ & $5\times10^{-5}$ \\
Warmup updates                    & 1,000 & 1,000 & 1,000 & 1,000 & 1,000 & 500 \\
AdamW $\beta_1$                   & 0.9 & 0.9 & 0.9 & 0.9 & 0.9 & 0.9 \\
AdamW $\beta_2$                   & 0.95 & 0.99 & 0.95 & 0.95 & 0.95 & 0.99 \\
Weight decay                      & 0.01 & 0.05 & 0.01 & 0.01 & 0.01 & 0.01 \\
Gradient clipping norm            & 1 & 1 & 1 & 1 & 1 & 1 \\
Codebook EMA decay                & 0.99 & 0.99 & 0.95 & 0.95 & 0.95 & 0.99 \\
Nested dropout                    & 0.5 & 0.7\,/\,0.3 & 0.5 & 0.5 & 0.5 & 0.5 \\
\bottomrule
\end{tabular}
\vspace{6pt}
\caption{\textbf{Tokenizers.} Text vocabulary size includes unused padding rows. Nested
dropout is the probability
of dropping the finest levels; LM1B trains on the complete and a truncated hierarchy,
weighted $0.7$ and $0.3$.}
\label{tab:tokenizer-config}
\label{tab:tokenizer-training}
\end{table}

\Needspace*{44\baselineskip}
\subsection{Generator (\name)}
\label{app:hparams}
\label{app:generator-training}
\label{app:code-targets}

\begin{table}[H]
\centering
\footnotesize
\setlength{\tabcolsep}{2.2pt}
\renewcommand{\arraystretch}{1.08}
\begin{tabularx}{\linewidth}{@{}>{\raggedright\arraybackslash}p{3.4cm} *{6}{>{\centering\arraybackslash}X}@{}}
\toprule
\textbf{Setting} & \textbf{OWT} & \textbf{LM1B} & \textbf{TinyStories} & \textbf{XSum} & \textbf{WMT14} & \textbf{Sudoku} \\
\midrule
\cfgrow{Architecture}
Parameters                        & 94.65M & 97.07M & 88.61M & 88.02M & 119.96M & 25.46M \\
Transformer blocks                & 12 & 12 & 12 & 12 & 12 & 8 \\
Hidden width                      & 768 & 768 & 768 & 768 & 768 & 512 \\
Attention heads                   & 12 & 12 & 12 & 12 & 12 & 8 \\
Prefix length $P$                 & 1 & 1 & 1 & 512 & 96 & 91 \\
RoPE base                         & $10^5$ & $10^4$ & $10^5$ & $10^5$ & $10^5$ & $10^4$ \\
Dropout                           & 0.1 & 0.1 & 0.1 & 0.3 & 0.1 & 0.1 \\
\midrule
\cfgrow{Training}
Training updates                  & 250,000 & 100,000 & 50,000 & 350,000 & 150,000 & 50,000 \\
Global batch size                 & 256 & 512 & 2,048 & 512 & 2,048 & 1,024 \\
Peak learning rate                & $3\times10^{-4}$ & $3\times10^{-4}$ & $3\times10^{-4}$ & $3\times10^{-4}$ & $3\times10^{-4}$ & $3\times10^{-4}$ \\
Learning-rate floor               & $3\times10^{-5}$ & $3\times10^{-5}$ & $3\times10^{-5}$ & $3\times10^{-5}$ & $3\times10^{-5}$ & $3\times10^{-5}$ \\
Warmup updates                    & 500 & 1,000 & 500 & 500 & 500 & 1,000 \\
AdamW $\beta_1$                   & 0.9 & 0.9 & 0.9 & 0.9 & 0.9 & 0.9 \\
AdamW $\beta_2$                   & 0.95 & 0.95 & 0.95 & 0.95 & 0.95 & 0.99 \\
Weight decay                      & 0.1 & 0.1 & 0.1 & 0.1 & 0.01 & 0.01 \\
Gradient clipping norm            & 1 & 1 & 1 & 1 & 1 & 1 \\
\bottomrule
\end{tabularx}
\vspace{6pt}
\caption{\textbf{Generators.} The prefix holds the source (XSum, WMT14), the puzzle
(Sudoku), or a null token. Training settings for XSum and WMT14 describe conditional
training. ``--'' means not applicable.}
\label{tab:generator-config}
\label{tab:generator-training}
\end{table}

\noindent\textbf{Training-token accounting.}\label{app:training-token-accounting}
Budgets count full-length text positions, including padding and repeated data exposure.
For LM1B, tokenizer and generator training total
$25{,}957\times2{,}048\times128+100{,}000\times512\times128
\approx13.36$B positions ($6.80$B+$6.55$B).
For OpenWebText, the corresponding total is
$78{,}432\times768\times1{,}024+250{,}000\times256\times1{,}024
\approx127.22$B positions ($61.68$B+$65.54$B), approximately $4\times$ fewer than
the $500$B-plus tokens used by MDLM and DUO
\citep{sahoo2024mdlm,sahoo2025duo}.
Scaling plots exclude the shared tokenizer. These budgets measure token exposure, not FLOPs.

Each generator block contains self-attention followed by a feed-forward layer.
The resized reconstruction $\hat z^{(k-1)}$ and prompt embeddings are projected to
the hidden width before the first block. Attention heads have dimension $64$,
equal to the hidden width divided by the number of heads.

\section{Baselines}
\label{app:baselines}

This appendix states where every baseline number in Section~\ref{sec:results} comes
from and how the baselines were sampled.

\subsection{Unconditional Generation}
\label{app:baselines-uncond}

\paragraph{LM1B.}
All LM1B baselines (RDLM, CANDI, MDLM, DUO, and FLM) use the checkpoints of
\citet{lee2026flm}, who trained them for the same $1$M iterations on $66$B tokens. We
sample each with $T\in\{8,12,16,32,64,128\}$ steps under the protocol of
Section~\ref{subsec:uncond}. RDLM is omitted from Figure~\ref{fig:gen-ppl} because its
Gen.\ PPL lies far above the other curves.

\paragraph{TinyStories.}
We train the AR, DUO, MDLM, and CANDI baselines ourselves using the same
transformer architecture and training settings as the TinyStories \name\ generator
(Table~\ref{tab:generator-config}). For FMLM+, we use the publicly released
checkpoint~\citep{agarwal2026posterior}.

\paragraph{OpenWebText.}
The MDLM, DUO, SEDD, and Eso-LM baselines use publicly available
checkpoints~\citep{sahoo2024mdlm,sahoo2025duo,lou2024sedd,sahoo2025esoteric}.
We sample each with $T\in\{8,16,\ldots,1{,}024\}$ steps under the same protocol.

\subsection{Conditional Generation}
\label{app:conditional}

The MDLM, DUO, E2D2, and SeqDiffuSeq results of Table~\ref{tab:seq2seq}
are taken from \citet{hu2026elf} (marked $^\dagger$), with their
configurations. We use publicly available checkpoints of ELF (with and without
guidance) and AURORA-LM~\citep{liang2026aurora} to reproduce their quality results
and measure sampling time under the setup of Section~\ref{subsec:efficiency}.
For conditional baselines without released checkpoints, we benchmark sampling
time using randomly initialized weights with the same model architecture and
task-specific sampling configuration reported for those tasks. These runs are
used only for timing; their quality results are taken from the published results.

\subsection{Reasoning}
\label{app:baselines-reasoning}

We follow the Sudoku evaluation setup of S-FLM~\citep{deschenaux2026language},
reporting the percentage of completely solved boards at three difficulty levels:
easy, medium, and hard, with $40$, $35$, and $30$ provided clues, respectively.
The baseline results in Table~\ref{tab:reasoning} are taken from
\citet{agarwal2026posterior}.

\section{Additional Results}
\label{app:extended_results}

This appendix checks that the LM1B comparison does not depend on the scoring model
(Appendix~\ref{app:ref-models}) and collects unconditional generation results from
Section~\ref{sec:results} (Appendix~\ref{app:sweeps}).

\subsection{Generative Perplexity Under Other Reference Models}
\label{app:ref-models}

To check whether the LM1B comparison depends on the reference model, we evaluate the same checkpoints and sampling settings as Figure~\ref{fig:gen-ppl} under four additional autoregressive models. Table~\ref{tab:gen-ppl-full} averages five sampling seeds with $256$ samples each. \name has the lowest average Gen.\ PPL, using approximately $5\times$ fewer training tokens and $12$ sampling steps rather than $128$.

\begin{table}[H]
\centering
\small
\setlength{\tabcolsep}{5pt}
\renewcommand{\arraystretch}{1.05}
\newcommand{\gh}[1]{{\scriptsize\textbf{#1}}}
\begin{tabular}{@{}l c c ccccc@{}}
\toprule
\gh{Model}
  & \gh{Size}
  & \gh{Tokens}
  & \gh{Mistral-7B\,$\downarrow$}
  & \gh{Llama-3.1-8B\,$\downarrow$}
  & \gh{Gemma-2-9B\,$\downarrow$}
  & \gh{Qwen-3-14B\,$\downarrow$}
  & \gh{Avg.\,$\downarrow$} \\
\midrule
RDLM  & 108M & 66B & 179.54 & 315.40 & 1293.67 & 379.00 & 541.90 \\
CANDI & 139M & 66B & 67.75  & 99.96  & 208.59  & 125.87 & 125.54 \\
MDLM  & 139M & 66B & 63.63  & 91.84  & 186.53  & 115.25 & 114.31 \\
DUO   & 139M & 66B & \textbf{55.08} & \underline{77.24} & \underline{141.16} & \underline{96.59} & \underline{92.52} \\
FLM   & 139M & 66B & 68.08  & 89.20  & 150.11  & 110.98 & 104.59 \\
\midrule
\rowcolor{oursrow}
\name & 97M & 13.36B & \underline{59.69} & \textbf{75.23} & \textbf{106.41} & \textbf{91.64} & \textbf{83.24} \\
\bottomrule
\end{tabular}
\vspace{6pt}
\caption{\textbf{LM1B Gen.\ PPL} (lower is better). Baselines use 128 steps; \name uses 12.
Size: generator parameters; Tokens: training budget (including tokenizer training for \name); Avg.: unweighted mean.}
\label{tab:gen-ppl-full}
\end{table}

\clearpage
\subsection{Unconditional Generation}
\label{app:sweeps}

Table~\ref{tab:uncond-all} collects the unconditional results of all three datasets at
the largest NFE reported for each baseline. We report generative perplexity under
GPT-2-large alongside mean per-sample unigram entropy $H$ to assess sample quality
and diversity together. The ground-truth data are included as a reference for both
metrics.

\begin{table}[H]
\centering
\footnotesize
\setlength{\tabcolsep}{3.2pt}
\renewcommand{\arraystretch}{1.05}
\newcommand{\na}{\textcolor{black!35}{--}}
\newcommand{\gr}[1]{\textcolor{black!55}{#1}}
\begin{tabular}{@{}l l ccc ccc ccc@{}}
\toprule
& & \multicolumn{3}{c}{\textbf{TinyStories}} & \multicolumn{3}{c}{\textbf{LM1B}}
  & \multicolumn{3}{c}{\textbf{OpenWebText}} \\
\cmidrule(lr){3-5}\cmidrule(lr){6-8}\cmidrule(l){9-11}
\textbf{Type} & \textbf{Method}
  & NFE\,$\downarrow$ & Gen.\ PPL\,$\downarrow$ & $H$
  & NFE\,$\downarrow$ & Gen.\ PPL\,$\downarrow$ & $H$
  & NFE\,$\downarrow$ & Gen.\ PPL\,$\downarrow$ & $H$ \\
\midrule
\gr{--} & \gr{Data} & \gr{--} & \gr{9.67} & \gr{3.96} & \gr{--} & \gr{53.51} & \gr{4.31}
  & \gr{--} & \gr{14.15} & \gr{5.44} \\
\midrule
\multirow{2}{*}{AR} & Sample & 128 & 8.89 & 4.01 & \na & \na & \na & 1024 & 35.45 & 5.58 \\
                    & Greedy & 128 & 5.34 & 4.06 & \na & \na & \na & 1024 & 1.07 & 1.65 \\
\midrule
\multirow{4}{*}{Disc.} & MDLM   & 128 & 17.89 & 4.03 & 128 & 115.45 & 4.33 & 1024 & 105.15 & 5.63 \\
                       & Duo    & 128 & 19.63 & 4.01 & 128 & 97.30  & 4.30 & 1024 & 77.69  & 5.55 \\
                       & SEDD   & \na & \na & \na & \na & \na & \na & 1024 & 101.43 & 5.61 \\
                       & Eso-LM & \na & \na & \na & \na & \na & \na & 1024 & 131.96 & 5.65 \\
\midrule
\multirow{5}{*}{Cont.} & RDLM   & \na & \na & \na & 128 & 295.40 & 4.35 & \na & \na & \na \\
                       & CANDI  & 128 & 22.74 & 4.03 & 128 & 124.64 & 4.34 & 1024 & 143.13 & 5.71 \\
                       & FLM    & 128 & 57.50 & 4.17 & 128 & 112.33 & 4.33 & 1024 & 62.23 & 5.33 \\
                       & S-FLM  & 128 & 95.25 & 4.08 & \na & \na & \na & 1024 & 123.87 & 5.52 \\
                       & FMLM+  & 128 & 15.43 & 3.90 & \na & \na & \na & 128  & 66.60  & 5.21 \\
\midrule
\rowcolor{oursrow} Ours & \name & 12 & 15.41 & 3.98 & 12 & 93.05 & 4.31
  & 32 & 83.12 & 5.42 \\
\bottomrule
\end{tabular}
\vspace{6pt}
\caption{\textbf{Unconditional generation on TinyStories, LM1B, and OpenWebText.}
Gen.\ PPL under GPT-2-large and mean per-sample unigram entropy $H$; the Data row scores
ground-truth text; ``--'' means no result on that dataset. Lower Gen.\ PPL at an entropy below that of
the data, as for FLM and FMLM+ on OpenWebText, partly reflects reduced diversity.}
\label{tab:uncond-all}
\end{table}

\section{Ablations}
\label{app:ablations}

We ablate the key design choices in \name: the latent code hierarchy
(Appendix~\ref{app:ablations-hierarchy}), the NCM transformer
(Appendix~\ref{app:ablations-generator}), code corruption
(Appendix~\ref{app:corruption}), and sampling (Appendix~\ref{app:ablation-sampling}).
Unless otherwise stated, we run our ablations on TinyStories~\citep{eldan2023tinystories}.
Each experiment changes one design choice and keeps every other setting fixed.

\subsection{Latent Code Hierarchy}
\label{app:ablations-hierarchy}

\subsubsection{Number of Levels}
\label{app:num-levels}

The number of levels $K$ sets the number of generation steps. We compare ladders on
TinyStories with $K\in\{4,8,16,32\}$. Each uses a linear schedule starting at $1$ and
ending at $128$.

\paragraph{Setup.}
Each ladder has its own tokenizer with $4{,}096$ codes per level, trained for $40$k
steps. We then train an NCM transformer on top of each tokenizer for $10$k steps.
We do not use code corruption, so the ladder is the only meaningful difference between
runs. We compute Gen.\ PPL using $512$ generated samples and multiple sampling seeds,
with GPT-2-Large as the reference model. Reference text has Gen.\ PPL $9.67$ and
unigram entropy $H=3.96$.

\begin{figure}[ht]
\centering
\includegraphics[width=\linewidth]{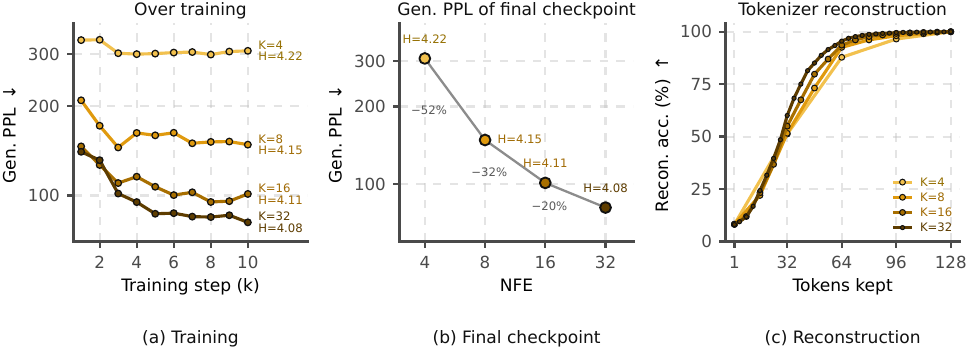}
\caption{\textbf{Ablation on the number of levels in the hierarchy.}
\textbf{(a)} Gen.\ PPL against training steps.
\textbf{(b)} Gen.\ PPL at the final checkpoint.
\textbf{(c)} Cumulative reconstruction accuracy per level.}
\label{fig:num-levels}
\end{figure}

Gen.\ PPL decreases as the number of levels in the hierarchy grows
(Figure~\ref{fig:num-levels}). It continues to improve at the largest hierarchy tested,
but with diminishing returns: further increases in the number of levels yield smaller
gains. Panel~(c) shows cumulative reconstruction accuracy; longer hierarchies
achieve slightly better final reconstruction, although all four reconstruct at least
$99.95\%$ of tokens. These ablations do not use code corruption, so errors can accumulate
across levels; adding corruption may narrow the gaps between hierarchies.

\FloatBarrier
\Needspace*{24\baselineskip}
\subsubsection{Level Schedule}
\label{app:hierarchy-shape}

The level schedule determines how the number of codes grows from coarse to fine.
We consider six schedules: linear, quadratic, square-root, logarithmic, exponential, and
cosine. Each has $K=12$ levels, starting at $1$ code and ending at $128$; code lengths are
rounded to integers. Figure~\ref{fig:hierarchy-shape} compares Gen.\ PPL over training,
Gen.\ PPL of the final checkpoint at $10$k steps, and the schedules themselves.

\begingroup
\setlength{\intextsep}{4pt}
\captionsetup{skip=4pt}
\begin{figure}[H]
\centering
\includegraphics[width=\linewidth]{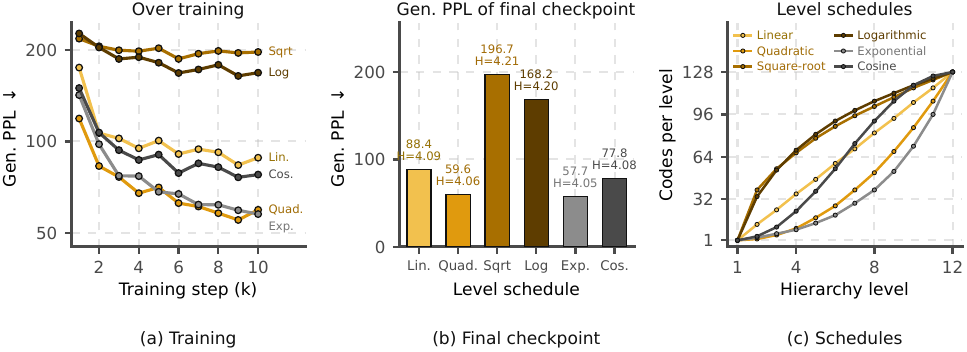}
\caption{\textbf{Ablation on the level schedule.}
\textbf{(a)} Gen.\ PPL against training steps.
\textbf{(b)} Gen.\ PPL of the final checkpoint at $10$k steps (bars).
\textbf{(c)} Code lengths for six level schedules.
$H$ is the final per-sample unigram entropy.}
\label{fig:hierarchy-shape}
\end{figure}
\endgroup

\FloatBarrier
\subsection{NCM Transformer}
\label{app:ablations-generator}

\Needspace*{16\baselineskip}
\subsubsection{Attention Pattern}
\label{app:attn-pattern}

For comparison purposes on TinyStories, we used the exact same $12$-level tokenizer,
generator, and selected conditioning-corruption recipe (mean rate $0.50$;
Appendix~\ref{app:corruption}) as in the previous example. The input blocks at each
level are aggregated from the earlier levels. Block-causal attention also includes
information from earlier blocks; a two-level window includes the preceding block.
Block-diagonal attention is the best option for generating low PPL and fast sampling
time with good preservation of entropy (Table~\ref{tab:attn-pattern}). We therefore
use this in \name.

\begin{table}[H]
\centering
\footnotesize
\setlength{\tabcolsep}{8pt}
\renewcommand{\arraystretch}{1.05}
\begin{tabular}{@{}l ccc@{}}
\toprule
\textbf{Attention} & \textbf{Gen.\ PPL} $\downarrow$ & $H$ & \textbf{Time (s)} $\downarrow$ \\
\midrule
Block-causal & 19.63 & 4.01 & 7.90 \\
Sliding window (2 levels) & 17.84 & 3.99 & 6.83 \\
Block-diagonal (ours) & \textbf{15.41} & 3.98 & \textbf{5.65} \\
\bottomrule
\end{tabular}
\vspace{3pt}
\caption{\textbf{Attention ablation on TinyStories.} Quality uses the selected corruption
recipe; $H$ is mean unigram entropy. Times are for 1{,}024 TinyStories samples at 12 steps
with batch size 16 (Section~\ref{subsec:efficiency}); block-causal and sliding-window times are
$1.40\times$ and $1.21\times$ the block-diagonal time.}
\label{tab:attn-pattern}
\end{table}

\begingroup
\setlength{\intextsep}{6pt}
\captionsetup{skip=4pt}
\begin{figure}[H]
    \centering
    \includegraphics[width=\linewidth]{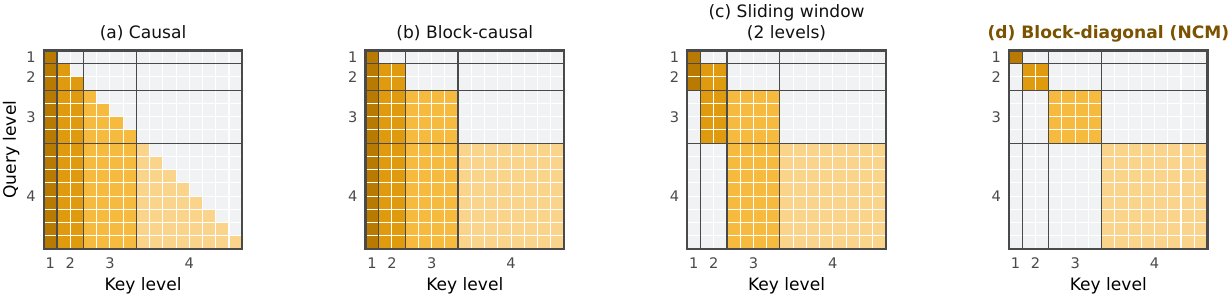}
    \caption{\textbf{Attention over cumulative-reconstruction blocks.} Allowed
    query--key pairs are colored by key level. \textbf{(a)}~Causal (reference).
    \textbf{(b)}~Block-causal. \textbf{(c)}~Two-level sliding window.
    \textbf{(d)}~Block-diagonal (\name): attention stays within the current block,
    whose input already summarizes earlier levels.}
    \label{fig:attn-patterns}
\end{figure}
\endgroup

\Needspace*{14\baselineskip}
\subsubsection{Context Encoder}
\label{app:text-encoder}

\par
\begingroup\setlength{\parfillskip}{0pt}\noindent
\begin{minipage}[t]{0.50\linewidth}
\vspace{0pt}
To support conditional generation, \name uses a prefix LM. One design decision is
how to encode the prefix. On XSum, we compare the frozen encoder from pretrained
T5-small with the encoder from our VQ-VAE (Table~\ref{tab:text-encoder}). T5-small
achieves slightly higher ROUGE scores: $+0.10$ ROUGE-1, $+0.15$ ROUGE-2, and
$+0.24$ ROUGE-L. However, it adds $35.3$M parameters, approximately a $40\%$
increase over our $88$M model, and introduces a dependency on an externally
pretrained language model. We therefore use the encoder from our VQ-VAE in all
reported experiments.
\end{minipage}\hfill
\begin{minipage}[t]{0.47\linewidth}
\vspace{0pt}
\centering
\footnotesize
\setlength{\tabcolsep}{3pt}
\renewcommand{\arraystretch}{1.05}
\begin{tabular}{@{}l c c c@{}}
\toprule
& \textbf{T5-small} & \textbf{Ours} & \\
\textbf{Metric} & \textbf{encoder} & \textbf{(VQ-VAE)} & \textbf{$\Delta$} \\
\midrule
ROUGE-1\,$\uparrow$          & 34.89  & 34.79 & $+0.10$ \\
ROUGE-2\,$\uparrow$          & 12.92  & 12.77 & $+0.15$ \\
ROUGE-L\,$\uparrow$          & 28.19  & 27.95 & $+0.24$ \\
\midrule
Parameters                   & 123.4M & \textbf{88.0M} & $+35.3$M \\
\bottomrule
\end{tabular}
\vspace{3pt}
\captionof{table}{\textbf{Context encoder: frozen T5-small vs.\ our VQ-VAE encoder.} Both use
the constant-LR recipe and greedy decoding at $8$ NFE, on the full XSum test set.}
\label{tab:text-encoder}
\end{minipage}
\par\endgroup

\subsubsection{Loss Weighting}
\label{app:loss-weighting}

\begin{figure}[!htbp]
\centering
\includegraphics[width=\linewidth]{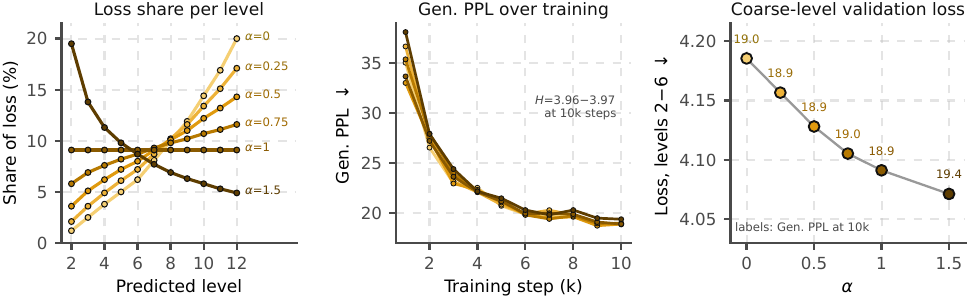}
\caption{\textbf{Loss weighting} on TinyStories. \textbf{Left:} loss share per level.
\textbf{Middle:} Gen.\ PPL during training. \textbf{Right:} coarse-level validation loss,
labelled with Gen.\ PPL after $10$k steps.}
\label{fig:loss-weighting}
\end{figure}

The loss weights level $k$ by $w_k\propto\ell_k^{-\alpha}$: $\alpha=0$ weights every code
equally, as the ELBO does (Appendix~\ref{app:theory:elbo}), and $\alpha=1$ every level
equally. On TinyStories (setup of Appendix~\ref{app:ts-setup}), raising $\alpha$ from $0$ to
$1.5$ moves the coarse levels' share of the loss from $19\%$ to $63\%$ and lowers their
validation loss, yet Gen.\ PPL after $10$k steps stays at $18.9$--$19.0$ for $\alpha\le1$
and rises slightly to $19.4$ at $1.5$ (Figure~\ref{fig:loss-weighting}).

\FloatBarrier
\subsection{Code Corruption}
\label{app:corruption}
\label{app:ts-setup}

\begin{wrapfigure}[14]{r}{0.39\linewidth}
\vspace{-\intextsep}
\centering
\includegraphics{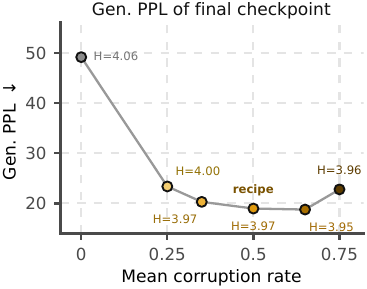}
\captionsetup{font=footnotesize,skip=4pt,justification=justified,singlelinecheck=false}
\caption{\textbf{Effect of code corruption.} Gen.\ PPL after $10$k steps vs.\
mean corruption rate. Nonzero rates use a shifted linear decreasing
schedule; $H$ is entropy.}
\label{fig:corruption}
\end{wrapfigure}
During training, \name uses teacher forcing and conditions on clean input codes;
during inference, it conditions on its own generated codes. This difference creates
opportunities for errors to accumulate across levels. Since codes within each level
are generated in parallel, multiple errors can enter the reconstruction used to
generate the next level. We address this mismatch by making the model resilient to
incorrect input codes: during training, we replace some input codes with entries
sampled uniformly from the same level's codebook, at a different rate per level.
Targets remain clean.

This ablation and the loss-weighting ablation (Appendix~\ref{app:loss-weighting}) use
TinyStories ($128$ tokens) and the $12$-level tokenizer of
Table~\ref{tab:tokenizer-config}. Each $88.6$M \name transformer is trained for
$10{,}000$ steps with a batch size of $1{,}024$. Gen.\ PPL is averaged over five
sampling seeds with $256$ samples each.
The degree of corruption matters: moderate corruption greatly improves generation
quality, while excessive corruption weakens the conditioning signal and degrades
quality (Figure~\ref{fig:corruption}). The type of
corruption, including how rates vary across levels and which replacement codes are
selected, also affects how closely training resembles generation. We use a linearly
decreasing corruption rate across levels. Corruption based on the model's actual
errors may better match inference conditions; we leave this to future work.

\FloatBarrier
\subsection{Sampling}
\label{app:ablation-sampling}
\label{app:sampling}

The quality of \name's generations may be enhanced during inference time without
retraining. The two types of decoding choices we examine in
Table~\ref{tab:seq2seq} are Classifier-Free Guidance and Minimum Bayes Risk (MBR)
Decoding.

\begin{figure}[h!]
\centering
\includegraphics[width=0.95\linewidth]{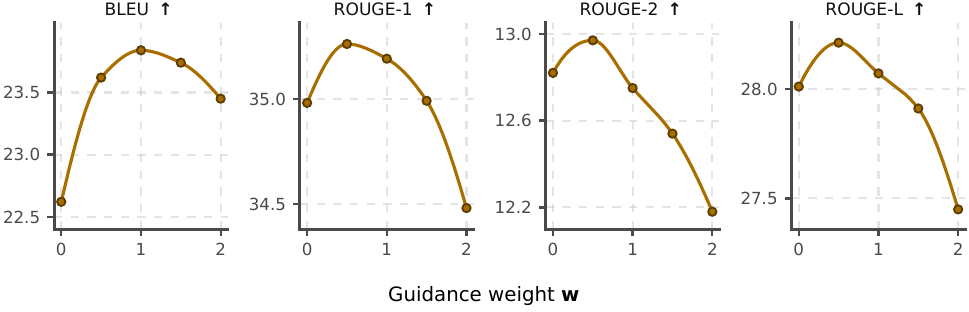}
\vspace{2pt}
\caption{\textbf{Effect of classifier-free guidance.} De--En BLEU and XSum
ROUGE-1/2/L as a function of the guidance weight $w$. All three ROUGE metrics
peak at $w\!=\!0.5$, the setting
reported in Table~\ref{tab:seq2seq}; BLEU, which is length-corrected, peaks later
at $w\!=\!1.0$.}
\label{fig:guidance}
\end{figure}

\noindent\textbf{Classifier-Free Guidance.}\label{app:guidance}
For Classifier-Free Guidance, we use VAR's approach \citep{tian2024var} to allow
for Classifier-Free Guidance in multiscale generations by combining conditional
and unconditional logits from each scale using a guidance weight $w$. As
illustrated in Figure~\ref{fig:guidance}, Classifier-Free Guidance results in less
improvement on XSum summarization compared to WMT14 translation.

\begin{figure}[h!]
\centering
\includegraphics[width=0.95\linewidth]{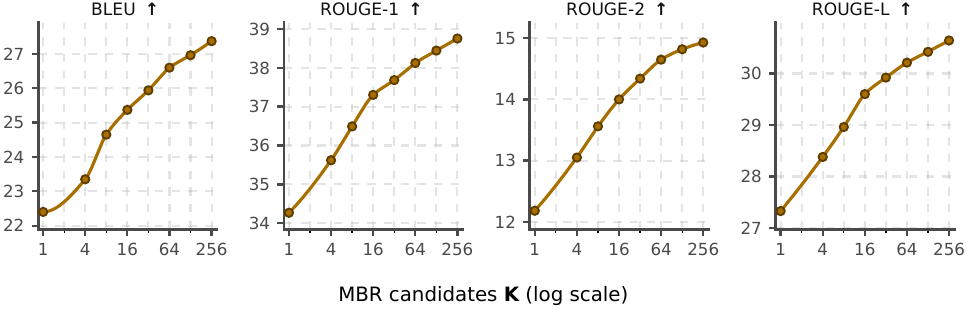}
\vspace{2pt}
\caption{\textbf{Effect of MBR candidate count.} De--En BLEU and XSum
ROUGE-1/2/L against the number of sampled candidates $K$, on a log-spaced axis.
Quality improves as more candidates are sampled. \name's low sampling cost allows
multiple candidates within the NFE budget that diffusion baselines use for a single
sample, enabling extra inference compute to be used for sampling and verification
through MBR selection.}
\label{fig:mbr}
\end{figure}

\noindent\textbf{Minimum Bayes Risk Decoding.}\label{app:mbr}
We draw $K$ candidates and keep the one with the highest average similarity to
the rest. Figure~\ref{fig:mbr} sweeps $K$: all three metrics rise monotonically
with diminishing returns, gaining $4.48$ ROUGE-1 in total from $K\!=\!1$ to
$K\!=\!256$.
The main-table settings use $K=16$ on XSum ($128$ NFE) and $K=12$ on WMT14
($132$ NFE), yielding $29.6$ ROUGE-L and $25.2$ BLEU, respectively.

\section{Tokenizer Analysis}
\label{app:analysis}
\label{app:vqvae}

\paragraph{Reconstruction \& Utilization.}
\label{app:tokenizer-levels}

\begin{figure}[ht]
  \centering
  \begin{minipage}[c]{0.32\linewidth}
    \centering
    \includegraphics[width=\linewidth]{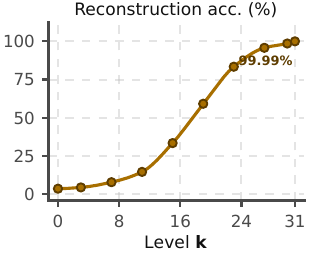}
  \end{minipage}\hfill
  \begin{minipage}[c]{0.32\linewidth}
    \centering
    \includegraphics[width=\linewidth]{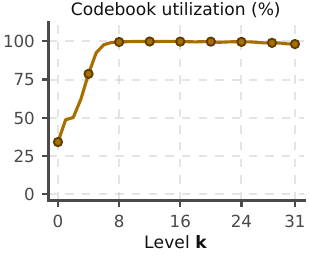}
  \end{minipage}\hfill
  \begin{minipage}[c]{0.32\linewidth}
    \centering
    \includegraphics[width=\linewidth]{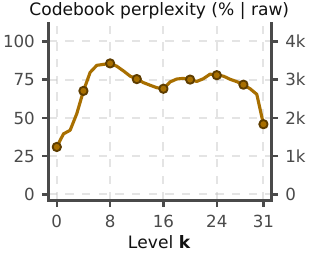}
  \end{minipage}
  \caption{\textbf{Reconstruction and utilization by level.} Left: exact token accuracy
  after decoding levels $0$ through $k$ at the final tokenizer checkpoint. Middle: percentage of the entries of each level's
  codebook selected at least once. Right: effective codebook utilization, measured by the
  perplexity of each level's code distribution, as a
  percentage of the codebook size.}
  \label{fig:tokenizer-levels}
\end{figure}

Figure~\ref{fig:tokenizer-levels} shows how the reconstruction builds up across the
hierarchy of the OpenWebText (OWT) tokenizer. After each level $k$, we decode the cumulative reconstruction $\hat z^{(k)}$
and measure exact token accuracy on held-out data. We also report the fraction of each
level's codebook entries used and its effective codebook utilization, measured by
codebook perplexity. Reconstruction accuracy increases across levels, reaching
$99.99\%$ at full resolution.

\paragraph{Tokenizer Failure Cases.}
\label{app:tokenizer-failure-cases}

We investigate typical tokenizer reconstruction errors using the OWT tokenizer as a
case study. As reported in Table~\ref{tab:recon}, it achieves $99.99\%$ exact token
reconstruction accuracy, corresponding to approximately $0.01\%$ incorrectly
reconstructed tokens. Table~\ref{tab:tok-failures} summarizes representative errors,
including substitutions of named entities and numerals, as well as errors in
diacritics, usernames, markup, and word boundaries.

\definecolor{tokerr}{HTML}{D03B3B}
\newcommand{\fx}[2]{{\color{black!45}\st{#1}}\,{\color{tokerr}\textbf{#2}}}

\begin{table}[ht]
\centering
\small
\renewcommand{\arraystretch}{1.25}
\begin{tabularx}{\linewidth}{@{}l >{\raggedright\arraybackslash}X@{}}
\toprule
\textbf{Category} & \textbf{Original} $\to$ \textbf{reconstruction}, in context \\
\midrule
Named entities   & \ldots{} City Metro \fx{Sydney}{Melbourne} Melbourne \fx{Brisbane}{Adelaide} Adelaide Perth \ldots{} \\
Diacritics       & \ldots{} ev sahibi \fx{tak{\i}m}{tak{\"a}m} skoru \ldots{} geriye \fx{d{\"u}{\c{s}}t{\"u}.}{d{\"u}{\c{s}}t{\"u}{\"u}} \ldots{} \\
Numerals         & \ldots{} 1957 \fx{1956}{1957} 1955 \fx{1954}{1955} 1953 \ldots{} 1949 \fx{1948}{1947} \fx{1947}{1946} 1946 \ldots{} \\
Usernames        & \ldots{} 10 maxion 10 fishingcat 10 \fx{da\_lan\_xiong\_mao}{da\_op\_xiong\_mao} \ldots{} \\
Markup           & \ldots{} \textless{}~head \fx{\textgreater{} \textless{}}{\textless{} \textgreater{}} title \fx{\textgreater{}}{\textless{}} \#BlackoutUK \ldots{} \\
Word boundaries  & \ldots{} Biological and Physical \fx{SciencesJournal}{Sciences Journal} of Chemical \ldots{} \\
\bottomrule
\end{tabularx}
\caption{\textbf{Tokenizer reconstruction errors} on OpenWebText, by category. Struck-through
grey: original; red: its reconstruction.}
\label{tab:tok-failures}
\end{table}

\clearpage
\section{Probing the Hierarchy}
\label{app:hierarchy}

We use the 12-level TinyStories model.
TinyStories~\citep{eldan2023tinystories} contains short, simple stories that make
changes easy to follow. We encode 128-token stories with the VQ-VAE to obtain
codes at each level of the hierarchy.

\noindent\textbf{Shared Structure, Different Realizations.}
We fix the first three levels of the hierarchy and randomly sample the remaining
nine levels using two seeds.
As shown in Figure~\ref{fig:story-continuation}, the outputs tend to preserve the
broad theme of the source story while varying names, objects, and wording.
Story A retains a girl seeking warmth and a garment that moves; Story B retains
Tim's alarm at something in the sky.

\begin{figure}[H]
\centering
\begingroup
\definecolor{ncmbranchyellow}{HTML}{FFF2A8}
\sethlcolor{ncmbranchyellow}
\newcommand{\ncmBranchChange}[1]{\hl{#1}}
\input{figs/story_continuation_samples}
\setlength{\parskip}{0pt}
\tcbset{ncmstorybranch/.style={
  enhanced, colback=white, colframe=black, arc=0pt, boxrule=0.5pt,
  left=1mm, right=1mm, top=0.5mm, bottom=0.5mm,
  before skip=0pt, after skip=0pt, fontupper=\fontsize{7.5}{8.2}\selectfont\raggedright}}

\begin{tcbraster}[raster columns=2, raster equal height=rows,
  raster column skip=6pt, raster before skip=0pt, raster after skip=3pt]
  \begin{tcolorbox}[ncmstorybranch]
    \textbf{Source story A}\hfill 128 tokens\par\vspace{1pt}
    \ncmBranchSourceA
  \end{tcolorbox}
  \begin{tcolorbox}[ncmstorybranch]
    \textbf{Source story B}\hfill 128 tokens\par\vspace{1pt}
    \ncmBranchSourceB
  \end{tcolorbox}
\end{tcbraster}

\begin{tcbraster}[raster columns=2, raster equal height=rows,
  raster column skip=6pt, raster row skip=2pt,
  raster before skip=0pt, raster after skip=3pt]
  \begin{tcolorbox}[ncmstorybranch]
    \textbf{Story A, seed 1}\hfill\ncmBranchAEarlyOneChanged\ changed\par\vspace{1pt}\ncmBranchAEarlyOne
  \end{tcolorbox}
  \begin{tcolorbox}[ncmstorybranch]
    \textbf{Story B, seed 1}\hfill\ncmBranchBEarlyOneChanged\ changed\par\vspace{1pt}\ncmBranchBEarlyOne
  \end{tcolorbox}
  \begin{tcolorbox}[ncmstorybranch]
    \textbf{Story A, seed 2}\hfill\ncmBranchAEarlyTwoChanged\ changed\par\vspace{1pt}\ncmBranchAEarlyTwo
  \end{tcolorbox}
  \begin{tcolorbox}[ncmstorybranch]
    \textbf{Story B, seed 2}\hfill\ncmBranchBEarlyTwoChanged\ changed\par\vspace{1pt}\ncmBranchBEarlyTwo
  \end{tcolorbox}
\end{tcbraster}

\endgroup
\vspace{4pt}
\captionsetup{font=small}
\caption{\textbf{Shared structure, different realizations.}
Each column shows one source story and two samples generated with its coarse
codes fixed. Text is shown
verbatim to the 128-token limit, including unfinished endings. Yellow highlights
added or replaced words relative to the corresponding source; deletions are not marked.}
\label{fig:story-continuation}
\end{figure}

\clearpage

\noindent\textbf{Mixing Coarse Representations from Different Stories.}
We encode pairs of 128-token stories from TinyStories with the VQ-VAE, mix their
coarse representations, and use NCM to generate the remaining finer levels
conditioned on the mixture. Figure~\ref{fig:story-composition} shows two source
pairs. Stories 1 and 2 blend drawing and billboard
details with the cold-weather and moving-object narrative. Stories 5 and 6 begin
with Tim and the jet before shifting to Max and the rain.

\begin{figure}[H]
\centering
\begingroup
\definecolor{ncmmixgreen}{HTML}{DDF1D4}
\definecolor{ncmmixyellow}{HTML}{FFF2A8}
\newcommand{\ncmMixFromA}[1]{{\sethlcolor{ncmmixgreen}\hl{#1}}}
\newcommand{\ncmMixFromB}[1]{{\sethlcolor{ncmmixyellow}\hl{#1}}}
\input{figs/story_composition_samples}
\setlength{\parskip}{0pt}
\tcbset{ncmstorymix/.style={
  enhanced, colback=white, colframe=black, arc=0pt, boxrule=0.5pt,
  left=1mm, right=1mm, top=0.5mm, bottom=0.5mm,
  before skip=0pt, after skip=0pt,
  fontupper=\fontsize{8}{9}\selectfont\raggedright}}

{\small
\begin{tabularx}{\linewidth}{@{}*{2}{>{\centering\arraybackslash}X}@{}}
\textbf{Stories 1 and 2} & \textbf{Stories 5 and 6}
\end{tabularx}\par}
\vspace{2pt}

\begin{tcbraster}[raster columns=2, raster equal height=rows,
  raster column skip=6pt, raster row skip=2pt,
  raster before skip=0pt, raster after skip=2pt]
  \begin{tcolorbox}[ncmstorymix]
    \textbf{\ncmMixFromA{Source A: Story 2}}\hfill 128 tokens\par\vspace{1pt}
    \ncmMixPairOneSourceA
  \end{tcolorbox}
  \begin{tcolorbox}[ncmstorymix]
    \textbf{\ncmMixFromA{Source A: Story 5}}\hfill 128 tokens\par\vspace{1pt}
    \ncmMixPairTwoSourceA
  \end{tcolorbox}
  \begin{tcolorbox}[ncmstorymix]
    \textbf{\ncmMixFromB{Source B: Story 1}}\hfill 128 tokens\par\vspace{1pt}
    \ncmMixPairOneSourceB
  \end{tcolorbox}
  \begin{tcolorbox}[ncmstorymix]
    \textbf{\ncmMixFromB{Source B: Story 6}}\hfill 128 tokens\par\vspace{1pt}
    \ncmMixPairTwoSourceB
  \end{tcolorbox}
\end{tcbraster}

\begin{tcbraster}[raster columns=2, raster equal height=rows,
  raster column skip=6pt, raster row skip=2pt,
  raster before skip=0pt, raster after skip=0pt]
  \begin{tcolorbox}[ncmstorymix]
    \textbf{Mixed coarse representations}\par\vspace{1pt}
    \ncmMixPairOneSeedOne
  \end{tcolorbox}
  \begin{tcolorbox}[ncmstorymix]
    \textbf{Mixed coarse representations}\par\vspace{1pt}
    \ncmMixPairTwoSeedOne
  \end{tcolorbox}
\end{tcbraster}
\endgroup
\vspace{2pt}
\captionsetup{font=small}
\caption{\textbf{Mixing coarse representations.}
Each column shows two sources and one output with newly generated finer levels.
Green (source A) and yellow (source B) highlight selected shared wording and
details. All text is verbatim, including unfinished endings.}
\label{fig:story-composition}
\end{figure}

\clearpage

\noindent\textbf{Perturbing Early Codes.}
We perturb a story's coarse representation by replacing two codes with random
codes observed at the same level across validation stories, then generate the
finer levels. Using the same sampling seed, we compare outputs with and without
these replacements. Figure~\ref{fig:story-perturbation} shows that changing two
coarse codes can alter characters, objects, and events while preserving parts of
the narrative.

\begin{figure}[H]
\centering
\begingroup
\definecolor{ncmperturbyellow}{HTML}{FFF2A8}
\sethlcolor{ncmperturbyellow}
\newcommand{\ncmPerturbChange}[1]{\hl{#1}}
\newcommand{\ncmPerturbCode}[2]{{\setlength{\fboxsep}{1pt}\colorbox{ncmperturbyellow}{$#1\!\to\!#2$}}}
\input{figs/story_perturbation_samples}
\setlength{\parskip}{0pt}
\tcbset{ncmstoryperturb/.style={
  enhanced, colback=white, colframe=black, arc=0pt, boxrule=0.5pt,
  left=1mm, right=1mm, top=0.5mm, bottom=0.5mm,
  before skip=0pt, after skip=0pt,
  fontupper=\fontsize{8}{9}\selectfont\raggedright}}

\begin{tcolorbox}[ncmstoryperturb]
  \textbf{Story 1: original validation text}\hfill 128 tokens\par\vspace{1pt}
  \ncmPerturbStoryOneOriginal
\end{tcolorbox}
\vspace{3pt}
{\scriptsize
\begin{tabularx}{\linewidth}{@{}*{8}{>{\centering\arraybackslash}X}@{}}
Position 0 & 1 & 2 & 3 & 4 & 5 & 6 & 7 \\
\ncmPerturbCode{3165}{512} & 3094 & \ncmPerturbCode{410}{2383} & 963 & 3438 & 370 & 3334 & 2301
\end{tabularx}\par}
\vspace{3pt}
\begin{tcbraster}[raster columns=2, raster equal height=rows,
  raster column skip=6pt, raster before skip=0pt, raster after skip=7pt]
  \begin{tcolorbox}[ncmstoryperturb]
    \textbf{Clean, seed 1}\par\vspace{1pt}
    \ncmPerturbStoryOneCleanOne
  \end{tcolorbox}
  \begin{tcolorbox}[ncmstoryperturb]
    \textbf{Perturbed, seed 1}\par\vspace{1pt}
    \ncmPerturbStoryOnePerturbedOne
  \end{tcolorbox}
\end{tcbraster}

\begin{tcolorbox}[ncmstoryperturb]
  \textbf{Story 2: original validation text}\hfill 128 tokens\par\vspace{1pt}
  \ncmPerturbStoryTwoOriginal
\end{tcolorbox}
\vspace{3pt}
{\scriptsize
\begin{tabularx}{\linewidth}{@{}*{8}{>{\centering\arraybackslash}X}@{}}
Position 0 & 1 & 2 & 3 & 4 & 5 & 6 & 7 \\
\ncmPerturbCode{1474}{219} & 3164 & 3742 & 2611 & \ncmPerturbCode{4055}{2225} & 1178 & 2580 & 3910
\end{tabularx}\par}
\vspace{3pt}
\begin{tcbraster}[raster columns=2, raster equal height=rows,
  raster column skip=6pt, raster before skip=0pt, raster after skip=0pt]
  \begin{tcolorbox}[ncmstoryperturb]
    \textbf{Clean, seed 1}\par\vspace{1pt}
    \ncmPerturbStoryTwoCleanOne
  \end{tcolorbox}
  \begin{tcolorbox}[ncmstoryperturb]
    \textbf{Perturbed, seed 1}\par\vspace{1pt}
    \ncmPerturbStoryTwoPerturbedOne
  \end{tcolorbox}
\end{tcbraster}
\endgroup
\vspace{3pt}
\captionsetup{font=small}
\caption{\textbf{Replacing two early codes.} Each example shows the original story,
its level-1 code replacements, and paired clean and perturbed outputs with the
same seed. Yellow marks word-level differences on both sides; punctuation-only
changes are unmarked. Code positions index eight equal regions of the 128-token
grid. Text is verbatim, including unfinished endings.}
\label{fig:story-perturbation}
\end{figure}

\clearpage

\section{Limitations \& Future Work}
\label{app:limits}

\name shows promising results on unconditional generation and downstream conditional
generation tasks such as summarization and translation. Several directions remain for
further improvement. \name fixes both the hierarchy depth $K$ and the per-level lengths
$\ell_k$, so even short continuations traverse the full hierarchy; an input-dependent
schedule would adapt compute to example complexity. \name relies on a separately
trained multiscale VQ-VAE tokenizer, decoupling representation learning from the
generative objective. An end-to-end training approach could benefit \name by aligning
representation learning with the generative objective. Our main experiments use generators of roughly $100$M
parameters, and our scaling study extends to $725$M parameters on LM1B. Scaling to
larger datasets and model sizes is another exciting direction. Generations are
generally fluent but exhibit occasional artifacts at coarse-level transitions,
presenting another opportunity to further improve \name.

\section{Additional Qualitative Samples}
\label{sec:qualitative-extra}

Figure~\ref{fig:samples} presents short samples from WMT14 German--English, XSum, LM1B, and OWT. In this section, we give three additional samples each for WMT14 German--English, XSum, TinyStories,
LM1B, and OpenWebText, with the relevant metrics reported where available,
and we follow \name\ through one hard Sudoku puzzle step by step (\S\ref{app:sudoku-levels}).

\subsection{WMT14 De--En}
\label{app:samples-wmt}

\begin{tcolorbox}[samplestyle, colback=white, colframe=black]
\scriptsize\linespread{1.06}\selectfont
\spheading{black}{Sample 1}{\textbf{BLEU:} 30.6}
\settowidth{\smplbl}{\textbf{Generated:}}\begin{tabular}{@{}l@{\hspace{6pt}}p{\dimexpr\linewidth-\smplbl-6pt\relax}@{}}
\splabel{black}{German}\textbf{:} &
  Volkswirtschaftler sorgen sich auch wegen der hohen
  Verbraucherverschuldung in Thailand, das diese Woche knapp aus einer
  technischen Rezession herausgekommen ist. \\
\end{tabular}
\sprule{black}
\sprows{black}{Economists are also worried about high levels of consumer debt
  in Thailand, which this week narrowly emerged from technical recession.}{Economists also worry about high consumer debt in Thailand, which has just
  come out of a technical recession this week.}
\end{tcolorbox}

\begin{tcolorbox}[samplestyle, colback=white, colframe=black]
\scriptsize\linespread{1.06}\selectfont
\spheading{black}{Sample 2}{\textbf{BLEU:} 55.7}
\settowidth{\smplbl}{\textbf{Generated:}}\begin{tabular}{@{}l@{\hspace{6pt}}p{\dimexpr\linewidth-\smplbl-6pt\relax}@{}}
\splabel{black}{German}\textbf{:} &
  Au\ss{}erdem erlebte er auf seiner Bahnfahrt von Titisee-Neustadt nach
  Freiburg die bereits unternommenen Anstrengungen f\"ur den Ausbau des
  \"offentlichen Personennahverkehrs. \\
\end{tabular}
\sprule{black}
\sprows{black}{In addition to this, on his train journey from Titisee-Neustadt to
  Freiburg, he experienced first-hand the efforts already undertaken to
  develop public transport.}{Furthermore, on his train journey from Titisee-Neustadt to Freiburg, he
  experienced the efforts already made to develop local public transport.}
\end{tcolorbox}

\begin{tcolorbox}[samplestyle, colback=white, colframe=black]
\scriptsize\linespread{1.06}\selectfont
\spheading{black}{Sample 3}{\textbf{BLEU:} 18.6}
\settowidth{\smplbl}{\textbf{Generated:}}\begin{tabular}{@{}l@{\hspace{6pt}}p{\dimexpr\linewidth-\smplbl-6pt\relax}@{}}
\splabel{black}{German}\textbf{:} &
  Samsung, Huawei und HTC stellen Handys her, die mit Googles Betriebssystem
  Android arbeiten, das in scharfem Wettbewerb zu den Mobilprodukten von
  Apple und Microsoft steht. \\
\end{tabular}
\sprule{black}
\sprows{black}{Samsung, Huawei and HTC all manufacture phones that operate on Google's
  Android operating system, which competes fiercely with Apple and Microsoft
  mobile products.}{Samsung, BMW and HTC make mobile phones that work with Android's operating
  operating system, which is firmly competing with Apple and and Microsoft
  products.}
\end{tcolorbox}

\subsection{Generation Trajectory}
\label{app:translation-trajectory}
Table~\ref{tab:translation-trajectory} shows the German source, reference, final
generated translation, and level-by-level outputs of a single generation run.
Figure~\ref{fig:generation-trajectory} shows
selected levels from the same run. Each generated row decodes the cumulative
reconstruction after that level. Intermediate quality is not monotonic.

\newcommand{\trbleu}[2][]{\makebox[1.6em][r]{#1#2}\hspace{4pt}\makebox[1.2cm][l]{\rlap{\color{black!8}\rule{1.2cm}{4pt}}{\color{cNCM}\rule{#2\dimexpr 0.012cm\relax}{4pt}}}}

\begin{table}[H]
\begin{tcolorbox}[samplestyle, colback=white, colframe=black]
\scriptsize\linespread{1.06}\selectfont
\spheading{black}{Translation trajectory}{\textbf{Final BLEU:} 77.8}
\settowidth{\smplbl}{\textbf{Generated:}}\begin{tabular}{@{}l@{\hspace{6pt}}p{\dimexpr\linewidth-\smplbl-6pt\relax}@{}}
\textbf{German:} &
  Es besteht allerdings die Sorge, dass kleine und mittlere Unternehmen angreifbar
  gegenüber Hacking und Überwachung bleiben. \\
\end{tabular}
\sprule{black}
\sprows{black}{{\tiny However, there is concern that small and medium-sized companies remain
  vulnerable to hacking and surveillance.}}{{\tiny However, there is concern that small and medium-sized enterprises will remain
  vulnerable to hacking and surveillance.}}
\sprule{black}
\renewcommand{\arraystretch}{1.3}\begin{tabularx}{\linewidth}{@{}c@{\hspace{10pt}}>{\raggedright\arraybackslash}X@{\hspace{10pt}}l@{}}
\textbf{Level} & \textbf{Decoded output after this level} & \textbf{BLEU}\,$\uparrow$ \\[-1pt]
\multicolumn{3}{@{}c@{}}{\color{black!35}\rule{\linewidth}{0.4pt}} \\[-5pt]
1 & However, it is a important to the the the and the. & \trbleu{6.0} \\
2 & However, it is a small to to small and and small, and small and and. & \trbleu{7.2} \\
3 & However, it is a concern for small small small medium-sized and-sized-sized and and and... & \trbleu{6.3} \\
4 & However, there is a medium of medium-sized enterprises remain remain remain remain to to to to. & \trbleu{16.0} \\
5 & However, there is a concern of medium medium-sized-sized enterprises remain remain remain and and orientation. & \trbleu{17.5} \\
6 & However, there is a concern that medium medium-sized enterprises will remain remain remain surveillance surveillance surveillance capabilities & \trbleu{17.7} \\
7 & However, there is that that and and medium-sized enterprises will remain remain vulnerable surveillance and surveillance. & \trbleu{24.2} \\
8 & However, there is that concern and and medium-sized enterprises will remain vulnerable to hacking and surveillance. & \trbleu{51.7} \\
9 & However, there is that that small and medium-sized enterprises will remain vulnerable to to and surveillance. & \trbleu{38.9} \\
10 & However, there is concern that small and medium-sized enterprises will remain vulnerable to to surveillance surveillance. & \trbleu{56.3} \\
\rowcolor{oursrow}
\textbf{11} & \textbf{However, there is concern that small and medium-sized enterprises will remain vulnerable to hacking and surveillance.} & \trbleu[\bfseries]{77.8} \\
\end{tabularx}
\end{tcolorbox}
\vspace{2pt}
\caption{\textbf{German--English translation trajectory.} Each row decodes the cumulative
reconstruction after one level of a single \name\ generation; text is shown as generated, and BLEU is computed per level against the reference. The final
level (shaded) is the output.}
\label{tab:translation-trajectory}
\end{table}

\subsection{XSum}
\label{app:samples-xsum}

\begin{tcolorbox}[samplestyle, colback=white, colframe=black]
\scriptsize\linespread{1.06}\selectfont
\spheading{black}{Sample 1}{\textbf{ROUGE-1:} 80.00\;\textcolor{black!25}{\textbar}\; \textbf{ROUGE-2:}
  71.43\;\textcolor{black!25}{\textbar}\; \textbf{ROUGE-L:} 80.00}
\noindent\splabel{black}{Context}\textbf{:}\;
The Brazil international, who had two years remaining on his current deal,
scored 13 goals in 59 games for the Blues last season. Willian, 27, joined
from Russian side Anzhi Makhachkala for a reported \pounds{}30m in 2013.
``It's a dream to continue with this club and over the next four years I will
try to help the team win more trophies,'' he said. Willian, who has scored
19 goals in 140 matches for Chelsea, was named player of the year last
season as the club finished 10th in the Premier League. He has won 30 caps
for Brazil since making his debut in 2011, scoring six goals. Last week,
Chelsea signed Belgium striker Michy Batshuayi from Marseille for a fee
believed to be worth \pounds{}33m, their first capture under new manager
Antonio Conte.
\sprule{black}
\sprows{black}{Chelsea midfielder Willian has signed a new four-year contract with the club.}{Chelsea midfielder Willian has signed a new four-year contract to stay with the Premier League champions.}
\end{tcolorbox}

\begin{tcolorbox}[samplestyle, colback=white, colframe=black]
\scriptsize\linespread{1.06}\selectfont
\spheading{black}{Sample 2}{\textbf{ROUGE-1:} 73.3\;\textcolor{black!25}{\textbar}\; \textbf{ROUGE-2:}
  57.1\;\textcolor{black!25}{\textbar}\; \textbf{ROUGE-L:} 73.3}
\noindent\splabel{black}{Context}\textbf{:}\;
The incident happened in Gerard Street, Ashton-in-Makerfield in Wigan at about
14:45 BST on Saturday. The woman, who has not been named but was in her 70s,
was critically injured and died at the scene. Gerard Street was closed for a
time while investigations were carried out. Police have appealed for any
witnesses to contact them.
\sprule{black}
\sprows{black}{A woman has died after being hit by a lorry in Greater Manchester, police
  have said.}{A woman has died after she was hit by a car in Greater Manchester.}
\end{tcolorbox}

\begin{tcolorbox}[samplestyle, colback=white, colframe=black]
\scriptsize\linespread{1.06}\selectfont
\spheading{black}{Sample 3}{\textbf{ROUGE-1:} 58.82\;\textcolor{black!25}{\textbar}\; \textbf{ROUGE-2:}
  12.50\;\textcolor{black!25}{\textbar}\; \textbf{ROUGE-L:} 41.18}
\noindent\splabel{black}{Context}\textbf{:}\;
The scene from Alice in Wonderland in Wardie Primary's assembly hall dates
back to 1936 and was recently restored with a Heritage Lottery Fund.

A mother has lodged a complaint about the image describing it as racist.

Edinburgh City Council said it understands the offensiveness of the image
but said it does not reflect the attitudes of the school.

An Edinburgh City Council spokesman said: ``The Alice in Wonderland mural
at Wardie Primary School was painted in 1936 and is of both historical and
artistic importance as evidenced by the fact it recently received full
Heritage Lottery Funding support to restore the work.

``While we understand the offensiveness of the image, it is in no way
indicative of the attitudes of either the school or the council.

``Our equalities policies and approaches are robustly multi-cultural and
anti-racist, promoting diversity and good relationships among pupils.''

A Police Scotland spokesman said: ``Police in Edinburgh have received a
complaint in relation to a mural at a primary school in the Trinity area.

``Officers are now liaising with Edinburgh City Council education department
with regards to this matter.

``Police Scotland treats all reports relating to hate incidents extremely
seriously and will thoroughly investigate whenever a report of this nature
is made.''
\sprule{black}
\sprows{black}{Police are investigating a complaint about a mural at an Edinburgh primary school which features a golliwog.}{A police investigation is under way after a mural was painted at a primary school in Edinburgh.}
\end{tcolorbox}

\Needspace*{14\baselineskip}
\subsection{TinyStories}
\label{app:samples-tinystories}

\begin{tcolorbox}[samplestyle, colback=white, colframe=black]
\scriptsize\linespread{1.06}\selectfont
\spheading{black}{Sample 1}{\textbf{Gen.\ PPL:} 10.2\;\textcolor{black!25}{\textbar}\; \textbf{$H$:} 3.98}
\noindent Once upon a time, there was a little boy named Tim. Tim loved to play with his yellow blocks. One day, he wanted to build a tower with his blocks. He was happy.
Tim's best friend, Tom, came over to play. Tom asked Tim, ``Can I play with your blocks?'' Tim said, ``No,, you cannot play with my blocks. But we can get us to play for you.''
They played and had fun. They put the yellow blocks and made an even bigger tower. It was tall and colorful. Tim and Tom had so much fun. They built the tower and made\ldots{}
\end{tcolorbox}

\begin{tcolorbox}[samplestyle, colback=white, colframe=black]
\scriptsize\linespread{1.06}\selectfont
\spheading{black}{Sample 2}{\textbf{Gen.\ PPL:} 14.1\;\textcolor{black!25}{\textbar}\; \textbf{$H$:} 3.88}
\noindent Once upon a time, there was a little dog named Sam. Sam was a big dog who liked to pull rope. He often liked to watch his pants spin and wagging his tail very much.
One day, Sam wanted his rope to his side, but the rope was too high for him to move. Sam felt sad and did not know what to do. His friend, Tom the cat, saw what was wrong. Sam asked him back for help.
Sam and Sam tried to pull the rope together. They pulled and pulled, but the rope would not move. Sam was sad and he could not get to his side.
\end{tcolorbox}

\begin{tcolorbox}[samplestyle, colback=white, colframe=black]
\scriptsize\linespread{1.06}\selectfont
\spheading{black}{Sample 3}{\textbf{Gen.\ PPL:} 17.9\;\textcolor{black!25}{\textbar}\; \textbf{$H$:} 4.03}
\noindent Once upon a time, there was a small brother named Sam. One day, Tim and Sam to the table. He saw something unusual in the kitchen. It was a little egg out of toast.
Tim said, ``Look, Sam! A little little egg!'' Sam looked and stared at the tree. Suddenly, it started to crack. The egg began to shake. Tim was scared. He did not want to lose the eggs or the cake. He watched as the little cake landed on the tree branch.
Tim picked it up and smiled. ``Look, Mom! The cake will fall back!'' he said. His mom came\ldots{}
\end{tcolorbox}

\subsection{LM1B}
\label{app:samples-lm1b}

\begin{tcolorbox}[samplestyle, colback=white, colframe=black]
\scriptsize\linespread{1.06}\selectfont
\spheading{black}{Sample 1}{\textbf{Gen.\ PPL:} 39.4\;\textcolor{black!25}{\textbar}\; \textbf{$H$:} 4.29}
\noindent {[}CLS] swine flu cases have been confirmed across the united states.
{[}CLS] there are actually more people that see it---since you
don\textquotesingle{}t worry about other things. {[}CLS] the labor department
says. {[}CLS] the two proposals are approved by the end of the week. {[}CLS] he
soon said that he always felt him about health care which was not a practical
issue and insisted on comprehensive reforms to improve a child\textquotesingle{}s
health. {[}CLS] the euro also added 1.33 per cent to \$1.46829 against the dollar
and gained 0.8 percent tob72,54 against the yen and slipped 2.4 percent barrel
to {[}CLS]
\end{tcolorbox}

\begin{tcolorbox}[samplestyle, colback=white, colframe=black]
\scriptsize\linespread{1.06}\selectfont
\spheading{black}{Sample 2}{\textbf{Gen.\ PPL:} 66.1\;\textcolor{black!25}{\textbar}\; \textbf{$H$:} 4.39}
\noindent {[}CLS] to be welcome---the united nations security resolution that was
imposed by russia earlier this month, the united states, european union and the
baltic states to take sanctions against iran, whose nuclear program also is its
west and nuclear programme. {[}CLS] they will cover the amount of state
subsidies on power and equipment. {[}CLS] any power cuts rates will be likely on those who
are still in talks over a deal late thursday, according to a government report
released late tuesday. {[}CLS] conservative advocates have done their effort to
helping people who have fallen before given to come into life while others has
in particular turn them off work, he said. {[}CLS] more than two {[}CLS]
\end{tcolorbox}

\begin{tcolorbox}[samplestyle, colback=white, colframe=black]
\scriptsize\linespread{1.06}\selectfont
\spheading{black}{Sample 3}{\textbf{Gen.\ PPL:} 72.0\;\textcolor{black!25}{\textbar}\; \textbf{$H$:} 4.35}
\noindent {[}CLS] that individuals should be given an perspective on where the
things are right now, and when we know things are doing well you know that the
federal government should still take measures to get us back over the next
three months. {[}CLS] \textquotedbl{}it is certainly saying, by 50 it
isn\textquotesingle{}t going to be put this week in charge, but this has been a
ridiculous prospect.. \textquotesingle{}i thought would say we have been very
good before election,\textquotedbl{} he added. {[}CLS] gs said the profit dropped 8 percent to $\pounds$96.7 million, and revenues grew 9 percent from \$10bn to
$\pounds$2.7m. {[}CLS] \textquotedbl{} {[}CLS]
\end{tcolorbox}

\Needspace*{22\baselineskip}
\subsection{OpenWebText}
\label{app:samples-owt}
We provide excerpts of $1{,}024$-token unconditional generations from our \name\ model
trained on OWT. Samples are shortened for space and readability; ellipses mark omissions.
Gen.\ PPL and entropy are computed on the full samples.

\begin{tcolorbox}[samplestyle, colback=white, colframe=black]
\scriptsize\linespread{1.06}\selectfont
\spheading{black}{Sample 1}{\textbf{Gen.\ PPL:} 47.8\;\textcolor{black!25}{\textbar}\; \textbf{$H$:} 5.40}
Compise of Doctor is the world's first shot villain in the Marvel Marvelion and appears to have host a full trailer for the beloved new Marvel franchise and Arrow in Daredevil. He has the upcoming show scenes in November 2015.

Welcome to the famous show, as well as those who joined Marvel films.

``Dark franchise is a wonderful, kind of exciting,'' Batman said in an interview. JoseRamos told reporters in a conference. ``I feel like I never happen, but never is a dream. I hope I do it.'' \ldots

``It's wonderful, wonderful actors,'' he said. ``You can't watch it and watch a video on screen.''

Watch the full video of Batman's future in Flash or Flash? \ldots

It's a difficult matter to imagine how it can make it easier for you to talk about content, since it's hard to work on what works.

Here's more information about the future, in that context of out aspects of ``wots,'', and the products that are merely cultural historical references to them online. Do you think the way they keep their web communication online? \ldots

Facebook's content is called ``The Internet,'' a web company called Media tool. It allows users to speed up interviews with some of the top web sites, including web sites, said Thomas William Abbartis, who just appeared on an online strategy that seemed to focus its attention to the web, its products and its groups. \ldots

Most of these companies are based on social sites, and Dr. Ababbartis's blog on Facebook's new site is to popularize web sites on its,{}, efficient manner. \ldots
\end{tcolorbox}

\begin{tcolorbox}[samplestyle, colback=white, colframe=black]
\scriptsize\linespread{1.06}\selectfont
\spheading{black}{Sample 2}{\textbf{Gen.\ PPL:} 26.4\;\textcolor{black!25}{\textbar}\; \textbf{$H$:} 5.27}
\ldots\ The growth is slowing down, the job creation is ramping up, and still again, a quarter of the Oregonians are stuck in part-time or without not work at all.

A new report shows Walmart's workforce gained economic growth this fall, dropping up an estimated 5,300 jobs last year alone, after increasing improvements for recovery. Instead, employment has declined, from 7.1 percent in August to to 7.3 percent in November, a five-year low.

But seasonally 272,195,000 people were unemployed in November, and just as many people were underemployed -{}- working part-time, but they could not get full-time jobs.

``Even though we're seeing [a] jobs recovery, there's still a lot of people who need to be working more than they are,'' said Nicke Belele, a Colorado statistician. \ldots

``It's just just about the quality of the jobs out there, it's the quality of jobs,'' State Department Mark McMullen said. ``We still have some work to do in terms of the quality of jobs.''

The November jobs report, issued Tuesday morning, will be the date of the last year. McMullen said it makes clear that Oregon has undergone rapid economic growth this fall, but the damage left by the workforce still can't be developed. The longer they are slightly underizedized, the bigger the focus on Oregon's economic growth potential. \ldots

And things could turn for things worse quickly. An estimated 17,000 Oregonians will run out of jobless unemployment in two weeks, when the long-term recovery emerged. \ldots
\end{tcolorbox}

\begin{tcolorbox}[samplestyle, colback=white, colframe=black]
\scriptsize\linespread{1.06}\selectfont
\spheading{black}{Sample 3}{\textbf{Gen.\ PPL:} 41.5\;\textcolor{black!25}{\textbar}\; \textbf{$H$:} 5.12}
Three Canadian music star Thomas Smith partnered with Graham Graham's Matthew Helons. \ldots

Steve Feich forward: I am going to start a year later this year, so are your thoughts have changed completely?

Since then, I understand that I've managed to grow up as a person as a decent person. \ldots

What are your thoughts on my career with last year?

It was so important. It was important for a whole lot. But I just can't do just any other than everyone. But I was for the whole comfortable feeling, and that's what I was happy. I wanted to do. \ldots

How can you change your career, the enthusiasm in your season, and how too much do you feel when you're in the team?

There's a lot of opportunities and a lot of opportunities. I know that I would be happy, and I know a lot that they're actually the ones, and the ones who do, but I don't say what I wanted to do to. \ldots

How was it hard for my professional career? Do you think you spend your life and present for your life?

Well, I thought it to be fairly unfair at risk of being college football. I have always been informed, busy, and playing while working hard and as a user. As a result, it's too hard to get in the right five hours and yourself being focused on your sweet school school at the football level. It's so much harder than the other player.

What do you have to do?

I have a whole moment at the good balance, for which I'm grateful for you for what's going into my life. \ldots
\end{tcolorbox}

\clearpage
\subsection{Sudoku}
\label{app:sudoku-levels}

\begin{figure}[H]
  \centering
  \input{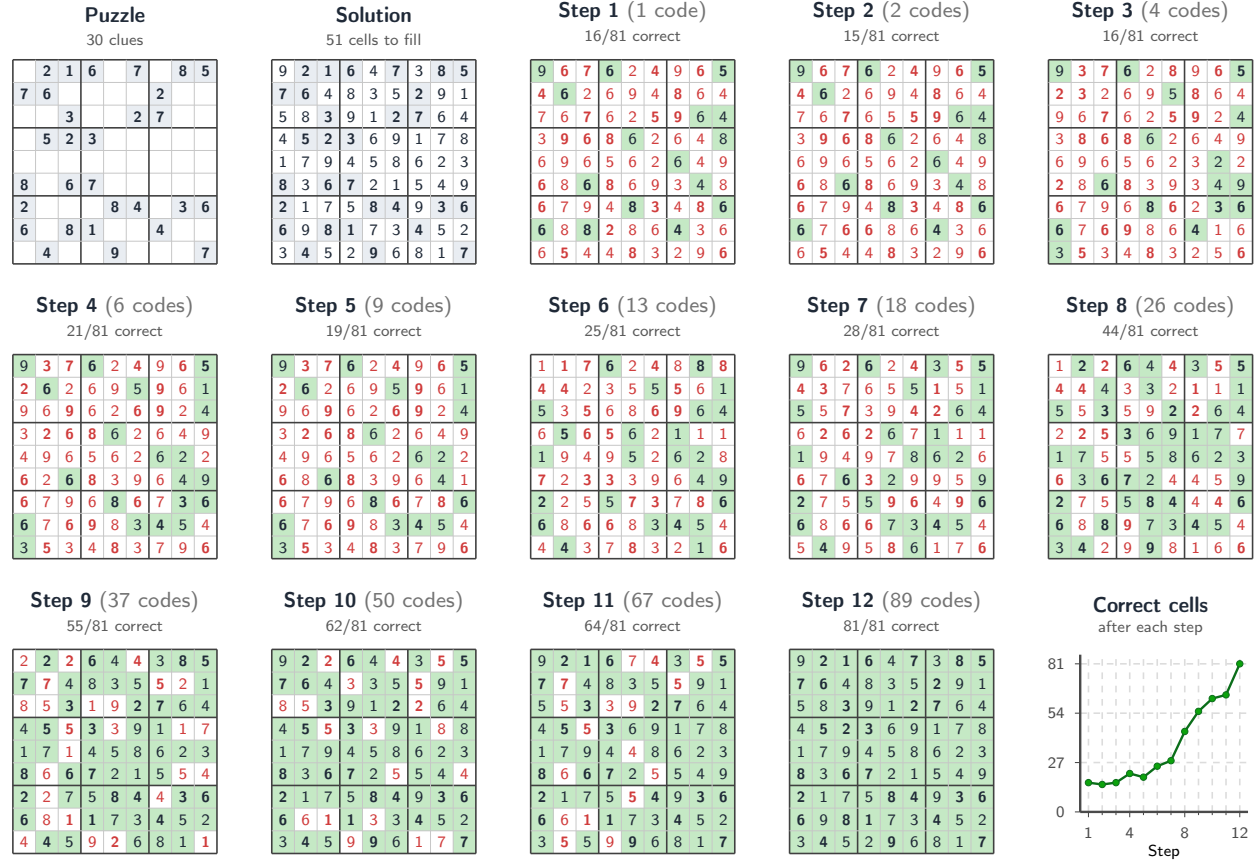}
  \caption{\textbf{Solving a hard Sudoku puzzle step by step.} Top left: the puzzle, with
  its 30 clues shaded, and its solution. Step $k$ decodes the codes that \name\ has
  generated in its first $k$ steps, where step $k$ generates level $k$ of the code
  hierarchy; its header gives the number of codes at that level and how many of the
  81 cells match the solution. Green cells are correct, red digits are wrong, and bold
  digits mark the clue positions. Bottom right: correct cells after each step.
  Across levels, \name\ iteratively refines the decoded board until it reaches the correct solution.}
  \label{fig:sudoku-hard}
\end{figure}

\end{document}